\documentclass[twocolumn,twoside,pdflatex]{IEEEtran}

\usepackage{amsmath}
\usepackage[pdftex]{color}
\usepackage[pdftex]{graphicx}
\usepackage[varg]{txfonts}
\usepackage{bm}
\usepackage{multirow}
\usepackage{url}
\usepackage{cite}
\usepackage{algorithm}
\usepackage{algorithmic}
\usepackage[right]{lineno}

\ifCLASSINFOpdf
\else
\fi
\newfont{\bg}{cmr9 scaled\magstep4}
\newcommand{\bigzerol}{\smash{\lower1.0ex\hbox{\bg 0}}}

\begin{document}

\title{
Unsupervised Detection of Anomalous Sound based on Deep Learning and the Neyman-Pearson Lemma
}

\author{
Yuma~Koizumi$^{1}$~\IEEEmembership{Member,~IEEE},
Shoichiro~Saito$^{1}$~\IEEEmembership{Member,~IEEE},
Hisashi~Uematsu$^{1}$~\IEEEmembership{Non-Member},
Yuta~Kawachi$^{1}$~\IEEEmembership{Non-Member},
and 
Noboru~Harada$^{1}$~\IEEEmembership{Senior Member,~IEEE}
\thanks{
All authors are with the NTT Media Intelligence Laboratories, NTT Corporation, Tokyo, Japan (e-mail: koizumi.yuma@ieee.org, \{saito.shoichiro, uematsu.hisashi, kawachi.yuta, noboru.harada\}@lab.ntt.co.jp).
A preliminary version of this work is published in \cite{Koizumi_2017_ADS}.

Copyright (c) 2018 IEEE. This article is the ``accepted'' version. Digital Object Identifier: 10.1109/TASLP.2018.2877258
}
}

\maketitle

\begin{abstract}
This paper proposes a novel optimization principle and its implementation for unsupervised anomaly detection in sound (ADS) using an autoencoder (AE). The goal of unsupervised-ADS is to detect unknown anomalous sound without training data of anomalous sound. Use of an AE as a normal model is a state-of-the-art technique for unsupervised-ADS. To decrease the false positive rate (FPR), the AE is trained to minimize the reconstruction error of normal sounds and the anomaly score is calculated as the reconstruction error of the observed sound. Unfortunately, since this training procedure does not take into account the anomaly score for anomalous sounds, the true positive rate (TPR) does not necessarily increase. In this study, we define an objective function based on the Neyman-Pearson lemma by considering ADS as a statistical hypothesis test. The proposed objective function trains the AE to maximize the TPR under an arbitrary low FPR condition. To calculate the TPR in the objective function, we consider that the set of anomalous sounds is the complementary set of normal sounds and simulate anomalous sounds by using a rejection sampling algorithm. Through experiments using synthetic data, we found that the proposed method improved the performance measures of ADS under low FPR conditions. In addition, we confirmed that the proposed method could detect anomalous sounds in real environments.
\end{abstract}



\section{Introduction}
\label{sec:intro}
\IEEEPARstart{A}{nomaly} detection in sound (ADS) has received much attention. 
Since anomalous sounds might indicate symptoms of mistakes or malicious activities, their prompt detection can possibly prevent such problems. 
In particular, ADS has been used for various purposes including audio surveillance \cite{Clavel_2005,Valenzise2007,Ntalampiras_2011,Foggia_2016}, animal husbandry \cite{Coucke_2003,Chung_2013_pig}, product inspection, and predictive maintenance \cite{Yamashita_2006,Koizumi_2017_ADS}.
For the last application, since anomalous sounds might indicate a fault in a piece of machinery, prompt detection of anomalies would decrease the number of defective product and/or prevent propagation of damage.
In this study, we investigated ADS for industrial equipment by focusing on machine-operating sounds.

ADS tasks can be broadly divided into supervised-ADS and unsupervised-ADS. 
The difference between the two categories is in the definition of anomalies. 
Supervised-ADS is the task of detecting ``{\it defined}'' anomalous sounds such as gunshots or screams \cite{Valenzise2007}, and it is a kind of rare sound event detection (SED) \cite{DCASE2017,rare_01,rare_02}. 
Since the anomalies are defined, we can collect a dataset of the target anomalous sounds even though the anomalies are rarer than normal sounds. 
Thus, the ADS system can be trained using a supervised method that is used in various SED tasks of the ``Detection and Classification of Acoustic Scenes and Events challenge'' (DCASE) such as audio scene classification \cite{task1_2016,task1_2017}, sound event detection \cite{task2_2016,task2_2017}, and audio tagging \cite{tagging}. 
On the other hand, unsupervised-ADS \cite{Hodge_2004,Patcha_2007,ASD_survey} is the task of detecting ``{\it unknown}'' anomalous sounds that have not been observed. 
In the case of real-world factories, from the view of the development cost, it is impracticable to deliberately be damaged the expensive target machine.
In addition, actual anomalous sounds occur rarely and have high variability. 
Therefore, it is impossible to collect an exhaustive set of anomalous sounds and need to detect anomalous sounds for which training data does not exist.
From this reason, the task is often tackled as the one-class unsupervised classification problem \cite{Hodge_2004,Patcha_2007,ASD_survey}.
This point is one of the major differences in premise between the DCASE tasks and ADS for industrial equipment.
Thus, in this study, we aim to detect unknown anomalous sounds based on an unsupervised approach.

In unsupervised anomaly detection, ``{\it anomaly}'' is defined as the patterns in data that do not conform to expected ``{\it normal}'' behavior \cite{ASD_survey}.
Namely, the universal set consists of only the normal and the anomaly, and the anomaly is the complement to the normal set.
More intuitively, 
the universal set is various machine sounds including many types of machines,
the normal set is one specific type of various machine sound, and 
the anomaly set is all other types of machine sounds.
Therefore, a typical way of unsupervised-ADS is the use of the outlier-detection technique. 
Here, the deviation between a normal model and an observed sound is calculated; the deviation is often called the ``{\it anomaly score}''. 
The normal model indicates the notion of normal behavior which is trained from training data of normal sounds. 
The observed sound is identified as an anomalous one when the anomaly score is higher than a pre-defined threshold value.
Namely, the anomalous sounds are defined as the sounds that do not exist in training data of normal sounds.

\begin{figure}[ttt]
\centering
\includegraphics[width=80mm]{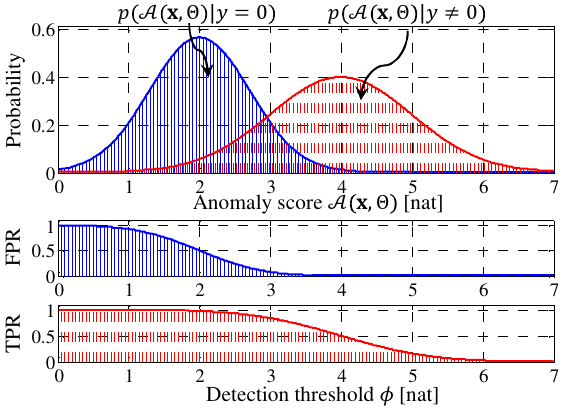}
\caption{
Trade-off relationship between anomaly score, true positive rate (TPR) and false positive rate (FPR). 
The top figure shows PDFs of anomaly scores for normal sounds (blue line) and anomalous sounds (red dashed line). 
The bottom figures show the FPR and TPR with respect to the threshold. 
When these PDFs overlap, a small threshold leads to a large TPR and FPR, and a large threshold leads to a small TPR and FPR. 
}
\label{fig:tprfpr}
\end{figure}

To train the normal model, it is necessary to define the optimality of the anomaly score. 
One of the popular performance measurements of ADS is to measure both the true positive rate (TPR) and false positive rate (FPR). 
The TPR is the proportion of anomalies that are correctly identified, and the FPR is the proportion of normal sounds that are incorrectly identified as anomalies. 
To improve the performance of ADS, we need to increase TPR and decrease FPR simultaneously. 
However, these metrics are related to the threshold value and have a trade-off relationship, as shown in Fig. \ref{fig:tprfpr}. 
When the PDFs of the anomaly scores of normal and anomalous sounds overlap, false detections cannot be avoided regardless of any threshold. 
Thus, to increase TPR and decreases FPR simultaneously, we need to train the normal model to reduce the overlap area. 
More intuitively, it is essential to provide small anomaly scores for normal sounds and large anomaly scores for anomalous sounds. 
In addition, if an ADS system gives a false alert frequently, we cannot trust it, just as ``{\it the boy who cried wolf }'' cannot be trusted. 
Therefore, it is especially important to increase TPR under a low FPR condition in a practical situation.

The early studies used various statistical models to calculate the anomaly score, such as the Gaussian mixture model (GMM) \cite{Ntalampiras_2011,Koizumi_2017_ADS} and support vector machine (SVM) \cite{Foggia_2016}. 
The recent literature calculates the anomaly score through the use of deep neural networks (DNN) such as 
the autoencoder (AE) \cite{Marchi_2015,Tagawa_2015,Marchi_2015_IJCNN,Kawaguchi_2017_MLSP} and variational AE (VAE) \cite{VAE_Anomaly,Kawachi_2018}. 
In the case of the AE, one is trained to minimize the reconstruction error of the normal training data, and the anomaly score is calculated as the reconstruction error of the observed sound. 
Thus, the AE provides small anomaly scores for normal sounds. 
However, it gives no guarantee to increase anomaly scores for anomalous sounds. 
Indeed, if the AE is generalized, the anomalous sounds will also be reconstructed and the anomaly score of anomalous sound will be small. 
Therefore, to increase TPR and decrease FPR simultaneously, the objective function should be modified.

Another strategy for unsupervised-ADS is the use of a generative adversarial network (GAN) \cite{GAN,VAEGAN}. 
GANs have been used to detect anomalies in medical images \cite{GAN_AD}. 
In this strategy, a generator simulates ``fake'' normal data, and a discriminator identifies whether the input data is a real normal data or not. 
Therefore, the discriminator can be trained to increase TPR for fake normal data and decrease FPR for true normal data simultaneously. 
However, since the generator is trained to make normal data, if it perfectly generates normal sounds, the anomaly score of normal sounds and FPR will increase. 
Therefore, it is necessary to build an algorithm to simulate ``non-normal'' sounds.

In this study, we propose a novel optimization principle and its implementation for ADS using AE. 
By considering an outlier-detection-based ADS as a statistical hypothesis test, we define optimality as an objective function based on the Neyman-Pearson lemma \cite{Ney_Pear}. 
The objective function works to increase TPR under an arbitrary low FPR condition. 
A problem in calculating TPR is the simulation of anomalous sound data. 
Here, we explicitly define the set of anomalous sounds to be the complement to the set of normal sounds and simulate anomalous sounds by using a rejection sampling algorithm. 

A preliminary version of this work is presented in \cite{Koizumi_2017_ADS}. 
The previous study utilized a DNN as a feature extractor, and the anomaly score was calculated using the negative-log-likelihood of a GMM trained from normal data. 
Thus, although the DNN was trained to maximize the objective function based on the Neyman-Pearson lemma, the normal model did not guarantee to increase TPR and decrease FPR. 
In this study, end-to-end training is achieved by using an AE as the normal model and both the feature extractor and the normal model are trained to increase TPR and decrease FPR. 

The rest of this paper is organized as follows. Section \ref{sec:Prev} briefly introduces outlier-detection-based ADS and its implementation using an AE. 
Section \ref{sec:Prop} describes the proposed training method and the details of the implementation. 
After reporting the results of objective experiments using synthetic data and verification experiments in real environments in Section \ref{sec:eval}, 
we conclude this paper in Section \ref{sec:conclusion}. 
The mathematical symbols are listed in Appendix \ref{sec:app_A}.

\section{Conventional method}
\label{sec:Prev}

\subsection{Identification of anomalous sound based on outlier detection}
ADS is an identification problem of determining whether the sound emitted from a target is a normal sound or an anomalous one. 
In this section, we briefly introduce the procedure of unsupervised-ADS.

First, an anomaly score $\mathcal{A} (\bm{\mathrm{x}}_{\tau}, \Theta)$ is calculated using a normal model. 
Here, $\bm{\mathrm{x}}_{\tau} \in \mathbb{R}^{\mathsf{Q}}$ is an input vector calculated from the observed sound indexed on $\tau \in \{ 1,2,...,T \}$ for time, and $\Theta$ is the set of parameters of the normal model. In many of the previous studies, $\bm{\mathrm{x}}_{\tau}$ was composed of hand-crafted acoustic features such as mel-frequency cepstrum coefficients (MFCCs) \cite{Clavel_2005,Valenzise2007,Ntalampiras_2011}, and the normal model was often constructed with a PDF of normal sounds. 
Accordingly, the anomaly score can be calculated as 
\begin{equation}
\mathcal{A}(\bm{\mathrm{x}}_{\tau}, \Theta) = - \ln p( \bm{\mathrm{x}}_{\tau} \mid \Theta, y=0),
\label{eq:log-likelihood}
\end{equation}
where $y$ denotes the state, $y=0$ is normal, and $y \neq 0$ is not normal, {\it i.e.} anomalous. 
$p(\bm{\mathrm{x}}|\Theta, y=0)$ is a normal model such as a GMM \cite{Koizumi_2017_ADS}. 
$\bm{\mathrm{x}}_{\tau}$ is determined to be anomalous when the anomaly score exceeds a pre-defined threshold value $\phi$:
\begin{equation}
\mathcal{H}(\bm{\mathrm{x}}_{\tau}, \Theta, \phi) =
 \begin{cases}
 0 \mbox{ }(\mbox{Normal}) & \mathcal{A} (\bm{\mathrm{x}}_{\tau}, \Theta) \leq \phi \\
 1 \mbox{ }(\mbox{Anomaly})& \mathcal{A} (\bm{\mathrm{x}}_{\tau}, \Theta) > \phi
 \end{cases}.
\label{eq:hard_thres}
\end{equation}
One of the performance measures of ADS consists of the pair of TPR and FPR. 
The TPR and FPR can be calculated as expectations of $\mathcal{H}(\bm{\mathrm{x}}, \Theta , \phi)$ with respect to anomalous and normal sounds, respectively:
\begin{align}
\mbox{TPR}(\Theta, \phi) &= 
\mathbb{E} \left[ \mathcal{H}(\bm{\mathrm{x}},\Theta , \phi) \right]_{\bm{\mathrm{x}} \mid y \neq 0}, \label{eq:TPR}\\
\mbox{FPR}(\Theta, \phi) &= 
\mathbb{E} \left[ \mathcal{H}(\bm{\mathrm{x}},\Theta , \phi) \right]_{\bm{\mathrm{x}} \mid y = 0} \label{eq:FPR},
\end{align}
where $\mathbb{E}[\cdot]_{x}$ denotes the expectation with respect to $x$. 
These metrics are related to $\phi$ and have a trade-off relationship as shown in Fig. \ref{fig:tprfpr}. 
The top figure shows the PDFs of anomaly scores for normal sounds $p(\mathcal{A} (\bm{\mathrm{x}}_{\tau}, \Theta) |y=0)$ and anomalous sounds $p(\mathcal{A} (\bm{\mathrm{x}}_{\tau}, \Theta) |y \neq 0)$.
 The bottom figures show the FPR and TPR with respect to $\phi$. 
When these PDFs overlap, false detections, {\it i.e.} false-positive and/or false-negative, cannot be avoided regardless of any $\phi$. 
In addition, the false detections increase as the overlap area gets wider. 
Therefore, to increase TPR and decrease FPR simultaneously, it is necessary to train $\Theta$ so that the anomaly score is small for normal sounds and large for anomalous sounds. 
More precisely, we need to train $\Theta$ to reduce the overlap area.

\subsection{Unsupervised-ADS using an autoencoder}
\label{sec:AE_ADS}

\begin{figure}[ttt]
\centering
\includegraphics[width=85mm]{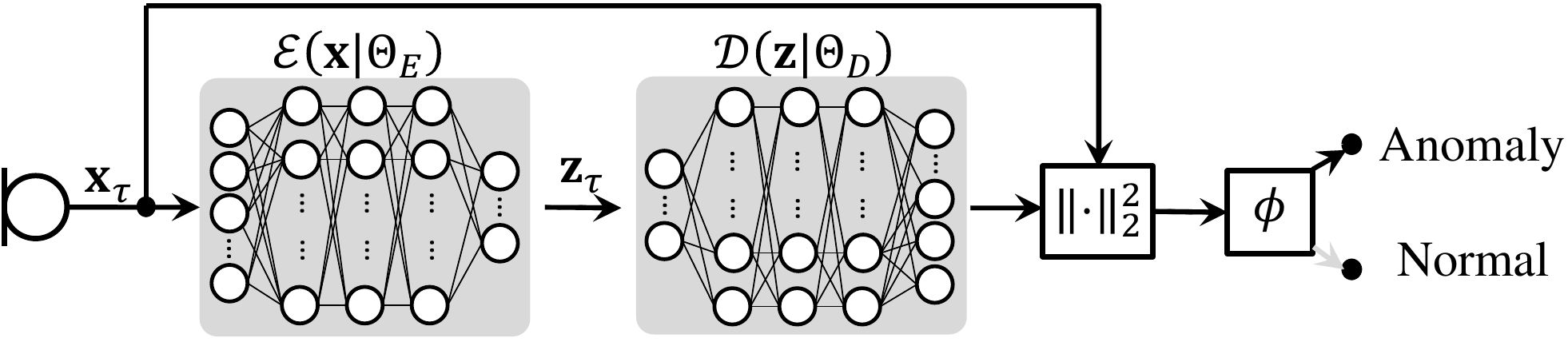}
\caption{
Anomaly detection procedure using autoencoder. 
The input vector is compressed and reconstructed by two networks $\mathcal{E}$ and $\mathcal{D}$, respectively. 
Since $\mathcal{E}$ and $\mathcal{D}$ are trained to minimize reconstruction error of normal sounds, the reconstruction error would be small if $\bm{\mathrm{x}}_{\tau}$ is normal. 
Thus, the anomaly score is calculated as a reconstruction error, and when the error exceeds a pre-defined threshold $\phi$, the observation is identified as anomalous.
}
\label{fig:basic_proc}
\end{figure}

Recently, deep learning has been used to construct a normal model. 
Several studies on deep-learning-based unsupervised-ADS have used an autoencoder (AE) \cite{Tagawa_2015,Marchi_2015,Marchi_2015_IJCNN,Kawaguchi_2017_MLSP}. 
This section briefly describes unsupervised-ADS using an AE (see Fig. \ref{fig:basic_proc}).

The goal of using an AE is to learn an efficient representation of the input vector by using two neural networks $\mathcal{E}$ and $\mathcal{D}$, which are called the encoder and decoder, respectively. 
First, the input vector $\bm{\mathrm{x}}$ is converted into a latent vector $\bm{\mathrm{z}} \in \mathbb{R}^{\mathsf{R}}$ by $\mathcal{E}$. 
Then, an input vector is reconstructed from $\bm{\mathrm{z}}$ by $\mathcal{D}$. 
These processes are expressed as
\begin{align}
\bm{\mathrm{z}} &= \mathcal{E}(\bm{\mathrm{x}} \mid \Theta _E), \label{eq;AE_encoder}\\
\hat{\bm{\mathrm{x}}} &= \mathcal{D}(\bm{\mathrm{z}} \mid \Theta _D). \label{eq;AE_decoder}
\end{align}
The parameters of both neural networks $\Theta = \{ \Theta _E , \Theta _D \}$ are trained to minimize the reconstruction error:
\begin{equation}
\mathcal{J}^{\mbox{\footnotesize AE}}(\Theta_E, \Theta_D) = 
\mathbb{E} \left[
\lVert \bm{\mathrm{x}} - \mathcal{D}(\mathcal{E}(\bm{\mathrm{x}} \mid \Theta _E) \mid \Theta _D ) \rVert_2 ^2 
\right]_{\bm{\mathrm{x}}}.
\label{eq:mmse_ae}
\end{equation}

In ADS using an AE, the anomaly score is the reconstruction error of the observed sound, which is calculated as
\begin{align}
\mathcal{A}(\bm{\mathrm{x}}_{\tau},\Theta) := \lVert \bm{\mathrm{x}}_{\tau} - \mathcal{D}(\mathcal{E}(\bm{\mathrm{x}}_{\tau} \mid \Theta _E) \mid \Theta _D ) \rVert_2 ^2.
\label{eq:anm_score_AE}
\end{align}
To train the normal model to provide small anomaly scores for normal sounds, the AE is trained to minimize the average reconstruction error of normal sound,
\begin{equation}
\mathcal{J}^{\mbox{\footnotesize AE}}(\Theta_E, \Theta_D) = 
\frac{1}{N^{(u)}}
\sum_{n=1} ^{N^{(u)}}
\mathcal{A}(\bm{\mathrm{x}}_n^{(u)},\Theta),
\label{eq:mmse_ae_sum}
\end{equation}
where $\bm{\mathrm{x}}_n^{(u)}$ is the $n$-th training data of normal sound and $N^{(u)}$ is the number of training samples of normal sound. 
This objective function works to decrease the anomaly score of normal sounds. 
However, there is no guarantee of increasing anomaly scores for anomalous sounds. 
Indeed, if the AE is generalized, the anomalous sounds will also be reconstructed and the anomaly score of anomalous sounds will be also small. 
Therefore, (\ref{eq:mmse_ae_sum}) does not ensure that false detections are reduced and the accuracy of ADS is improved; thus, it would be better to modify the objective function.

\section{Proposed method}
\label{sec:Prop}

We will begin by defining an objective function that builds upon the Neyman-Pearson lemma in Sec. \ref{sec:NeyPear}. 
Then, we will describe the rejection sampling algorithm for simulating anomalous sound used for calculating TPR in Sec \ref{sec:ae_NP}. 
After that, the overall training and detection procedure of the proposed method will be summarized in Sec. \ref{sec:full_procedure} and Sec. \ref{sec:full_procedure_detect}. 
As a modified implementation of proposed method, we extend the proposed method to an area under the receiver operating characteristic curve (AUC) maximization in Sec \ref{sec:NPextension}. 

\subsection{Objective function for anomaly detection based on the Neyman-Pearson lemma}
\label{sec:NeyPear}

From (\ref{eq:log-likelihood}) and (\ref{eq:hard_thres}), an anomalous sound satisfies the following inequality:
\begin{equation}
p(\bm{\mathrm{x}} \mid \Theta, y=0) < \exp (-\phi).
\label{eq:definition_anomaly}
\end{equation}
Since $\phi$ is assumed to be sufficiently large to avoid false positives, an anomalous sound can be defined as ``a sound which cannot be regarded as a sample of the normal model.'' 
Thus, we can regard outlier-detection-based ADS as a statistical hypothesis test. 
In other words, the observed sound is identified as anomalous when the following null hypothesis is rejected.

\vspace{5pt}
\noindent
{\bf Null hypotheses: }$\bm{\mathrm{x}}$ is a sample of the normal model $p(\bm{\mathrm{x}} \mid \Theta, y=0)$.

\vspace{5pt}

The Neyman-Pearson lemma \cite{Ney_Pear} states the condition for $\mathcal{A}(\bm{\mathrm{x}}, \Theta)$ that achieves the {\it most powerful test} between two simple hypotheses. 
According to it, the most powerful test has the greatest detection power among all possible tests of a given FPR \cite{MostPowerful}. 
More simply, the most powerful test maximizes the TPR under the constraint that the FPR equals $\rho$, {\it i.e.}, 
\begin{equation}
\nonumber
\mbox{ maximize } \mbox{TPR}(\Theta, \phi) \mbox{, subject to } \mbox{FPR}(\Theta, \phi) = \rho.
\end{equation}
Since the FPR can be controlled by manipulating $\phi$, we define $\phi_{\rho}$ as satisfying $\mbox{FPR}(\Theta, \phi_{\rho}) = \rho$. 
Accordingly, the objective function to obtain the most powerful test function can be defined as the one that maximizes $\mbox{TPR}(\Theta, \phi_{\rho})$ with respect to $\Theta$. 
However, since the FPR is also a function of $\Theta$, it may become large when focusing only on TPR. 
To maximize the TPR and minimize the FPR simultaneously, we train $\Theta$ to maximize the following objective function,
\begin{equation}
\mathcal{J}^{\mbox{\footnotesize NP}}(\Theta) = \mbox{TPR}(\Theta, \phi_{\rho}) - \mbox{FPR}(\Theta, \phi_{\rho}),
\label{eq:NP_obj_proposed}
\end{equation}
where the superscript ``NP'' is an abbreviation of ``Neyman-Pearson''.
Since the proposed objective function directly increases TPR and decreases FPR, $\Theta$ can be trained to provide a small anomaly score for normal sounds and a large anomaly score for anomalous sounds.

There are two problems when it comes to training $\Theta_E$ and $\Theta _D$ to maximize (\ref{eq:NP_obj_proposed}). 
The first problem is the calculation of TPR. 
The TPR and FPR are the expectations of $\mathcal{H}(\bm{\mathrm{x}},\Theta , \phi)$, and in most practical cases, the expectation is approximated as an average over the training data. 
Thus, to calculate TPR and FPR, we need to collect enough normal and anomalous sound data for the average to be an accurate approximation of the expectation. 
However, since anomalous sounds occur rarely and have high variability, this condition is difficult to satisfy. 
In section \ref{sec:ae_NP}, to calculate TPR, we consider ``anomaly'' to mean ``not normal'' and simulate anomalous sounds by using a sampling algorithm. 
The second problem is the determination of the threshold $\phi_{\rho}$. 
In a parametric hypothesis test such as a $t$-test, the threshold at which FPR equals $\rho$  can be analytically calculated. 
However, DNN is a non-parametric statistical model; thus, the threshold $\phi_{\rho}$ can not be analytically calculated. 
In section \ref{sec:full_procedure}, we numerically calculate $\phi_{\rho}$ as the $\lfloor \rho M \rfloor$-th value of the sorted anomaly scores of $M$ normal sounds, where $\lfloor \cdot \rfloor$ is the flooring function.

\subsection{Anomalous sound simulation using an autoencoder}
\label{sec:ae_NP}

\begin{figure}[ttt]
  \centering
  \includegraphics[width=85mm]{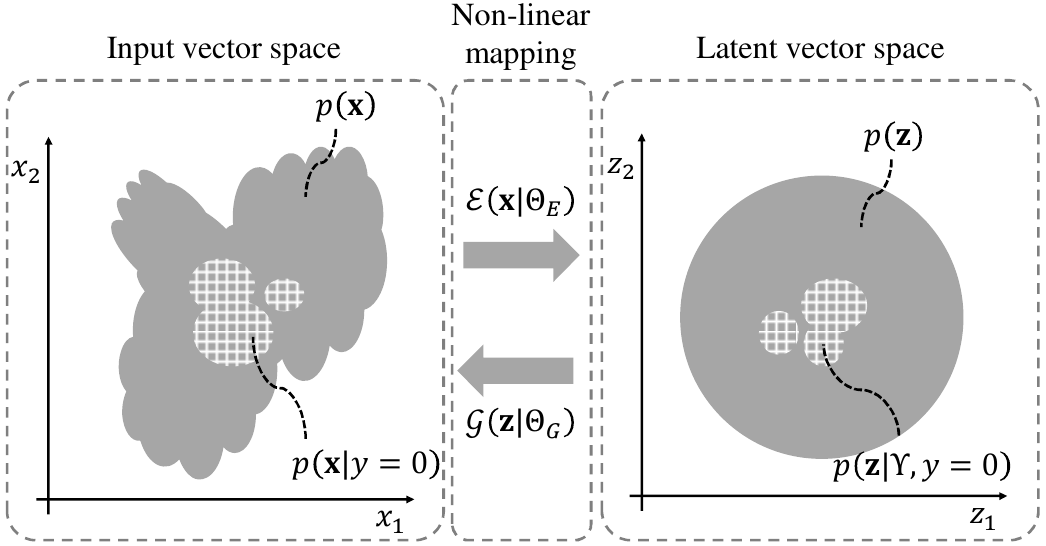}
  \caption{
Concept of PDFs of normal, various, and anomalous sounds using two neural networks. 
The PDF of normal sounds ({\it i.e.} meshed area) is a subset of the PDF of various sounds ({\it i.e.} gray area), and the PDF of anomalous sounds is expressed as complement of the PDF of normal sounds ({\it i.e.} inside the gray area and outside the meshed area). 
$\bm{\mathrm{x}}$ is mapped to $\bm{\mathrm{z}}$ by $\mathcal{E}$, and $\bm{\mathrm{z}}$ is reconstructed to $\tilde{\bm{\mathrm{x}}}$ by $\mathcal{G}$. 
Here, $\mathcal{E}$ and $\mathcal{G}$ are trained to satisfy $p(\bm{\mathrm{z}}) = \mathcal{N}( \bm{\mathrm{z}} | \bm{0}_{\mathsf{R}}, \bm{I}_{\mathsf{R}})$ and $\bm{\mathrm{x}} = \tilde{\bm{\mathrm{x}}}$, respectively. 
The PDF of the latent vector of normal sounds is modeled using a GMM $p(\bm{\mathrm{z}} \mid \Upsilon, y=0)$ given by (\ref{eq:prp:log-likelihood}). 
}
  \label{fig:anomalousPDF}
\end{figure}

\begin{figure*}[ttt]
  \centering
  \includegraphics[width=160mm]{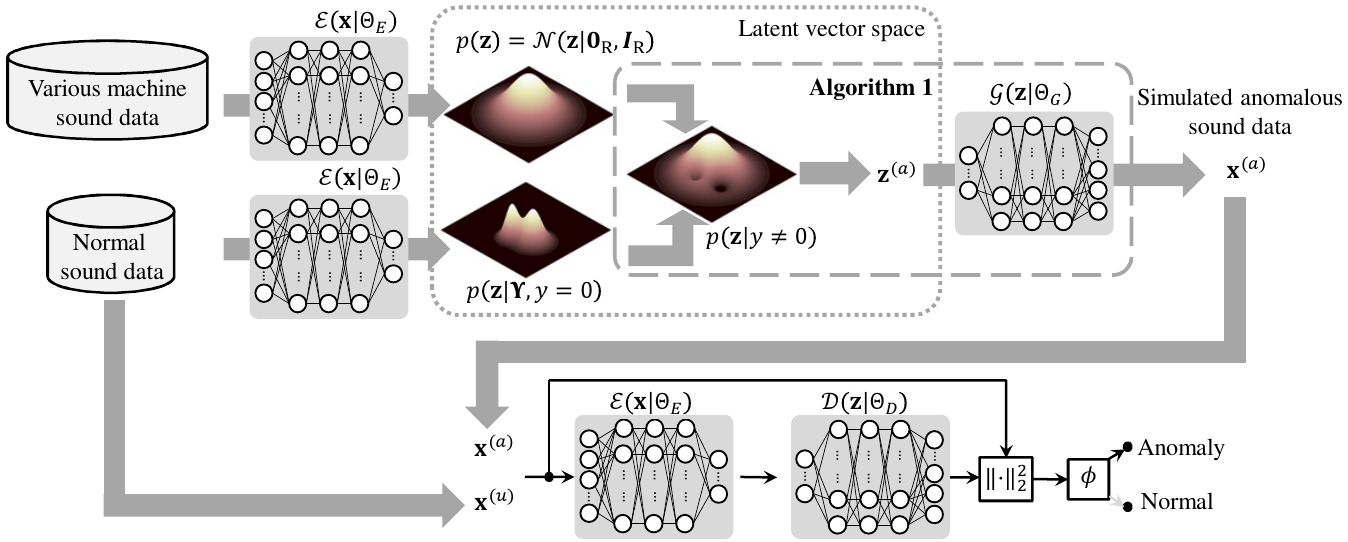}
  \caption{Procedure of anomalous sound simulation using autoencoder.}
  \label{fig:anomaly_simluation}
\end{figure*}

In accordance with (\ref{eq:definition_anomaly}), anomalous sounds emitted from the target machine are different from normal ones. 
Thus, we consider the set of normal sounds to be a subset of various machine sounds, and the set of anomalous sounds to be its complement. 
Then, we use rejection sampling to simulate anomalous sounds; namely, a sound is sampled from various machine-sound PDFs, and it is accepted as an anomalous sound when its anomaly score is high. 
However, since the PDF of various machine sounds in the input vector domain $p( \bm{\mathrm{x}} )$ may have a complex form, the PDF cannot be written in an analytical form and the sampling algorithm would become complex. 
Inspired by the strategy of VAE, we can avoid this problem by training $\mathcal{E}$ so that the PDF of various latent vectors $p(\bm{\mathrm{z}})$ is mapped to a PDF whose samples can be generated by a pseudorandom number generator from a uniform distribution and its variable conversion. 
Then, the latent vectors of anomalous sounds $\bm{\mathrm{z}}^{(a)}$ are sampled using the rejection sampling algorithm, and the input vectors of anomalous sounds $\bm{\mathrm{x}}^{(a)}$ are reconstructed using a third neural network $\mathcal{G}$,
\begin{align}
\bm{\mathrm{x}}^{(a)} = \mathcal{G}(\bm{\mathrm{z}}^{(a)} \mid \Theta _G),
\label{eq:prop_generator}
\end{align}
where $\Theta _G$ is the parameter of $\mathcal{G}$. 
Hereafter, we call $\mathcal{G}$ the generator. 
Although there is no constraint on the architecture of $\mathcal{G}$, we will use the same architecture for $\mathcal{D}$ and $\mathcal{G}$. 
In addition, to simply generate and reject a candidate latent vector, we use two constraints to train $\Theta_E$ and $\Theta_G$, and model the PDF of normal latent vectors using the GMM as 
\begin{align}
p(\bm{\mathrm{z}} \mid \Upsilon, y=0) &= \sum_{k=1}^{K} w_k \mathcal{N}(\bm{\mathrm{z}} \mid \bm{\mu}_k, \bm{\Sigma}_k) ,
\label{eq:prp:log-likelihood}
\end{align}
where $\Upsilon = \{w_k, \bm{\mu}_k, \bm{\Sigma}_k \mid k=1,...,K \}$, $K$ is the number of mixtures, and $w_k, \bm{\mu}_k$, and $\bm{\Sigma}_k$ are respectively the weight, mean vector, and covariance matrix of the $k$-th Gaussian. 
The concepts of these PDFs are shown in Fig. \ref{fig:anomalousPDF}, and the procedure of anomalous sound simulation is summarized in {\bf Algorithm \ref{alg:generate}} and Fig. \ref{fig:anomaly_simluation}.

\begin{algorithm}[tt] 
\caption{Simulation algorithm of anomalous sound in latent vector space.}
\label{alg:generate} 
\begin{algorithmic}[1]
\STATE {\bf Input: } Generator $\mathcal{G}$, GMM $p(\bm{\mathrm{z}} \mid \Upsilon, y=0)$ and $\phi_{z}$
\STATE $\ell \gets -\infty$
\WHILE{$\ell \leq \phi_{z}$}
\STATE Draw $\tilde{\bm{\mathrm{z}}}$ from $\mathcal{N}( \bm{\mathrm{z}} | \bm{0}_{\mathsf{R}} ,\bm{I}_{\mathsf{R}})$
\STATE Evaluate $\ell \gets - \ln p(\tilde{\bm{\mathrm{z}}} \mid \Upsilon, y=0)$
\ENDWHILE
\STATE $\bm{\mathrm{z}}^{(a)} \gets \tilde{\bm{\mathrm{z}}}$
\STATE Generate anomalous sound by $\bm{\mathrm{x}}^{(a)} = \mathcal{G}( \bm{\mathrm{z}}^{(a)} \mid \Theta _G)$
\STATE {\bf Output: } $\bm{\mathrm{x}}^{(a)}$
\end{algorithmic}
\end{algorithm}

First, we describe the two constraints for training $\Theta_E$ and $\Theta_G$. 
For algorithmic efficiency, $p(\bm{\mathrm{z}})$ should be generated with a low computational cost. 
As an implementation of $p(\bm{\mathrm{z}})$, we use the normalized Gaussian distribution, because its samples can be generated by a pseudorandom number generator such as the Mersenne-Twister. 
Thus, for training $\Theta_E$ and $\Theta_G$, we use the first constraint so that $\bm{\mathrm{z}}$ of the various machine sounds follows a normalized Gaussian distribution. To satisfy the first constraint, we train $\Theta_E$ to minimize the following Kullback-Leibler divergence (KLD):
\begin{align}
\nonumber
\mathcal{J}^{\mbox{\footnotesize KL}}(\Theta_{E}) 
&= D \left( \mathcal{N}(\bm{\mathrm{z}} \mid \bm{0}_\mathsf{R}, \bm{I}_\mathsf{R}) || \mathcal{N}(\bm{\mathrm{z}} \mid \bm{\mu}_{\mathcal{V}}, \bm{\Sigma}_{\mathcal{V}}) \right), \\
&= \frac{1}{2}
\left[ 
\ln | \bm{\Sigma}_{\mathcal{V}} |
 + \mbox{tr} \left\{ \bm{\Sigma}_{\mathcal{V}} ^{-1} \right\}
 + \bm{\mu}_{\mathcal{V}}^{\top} \bm{\Sigma}_{\mathcal{V}} ^{-1} \bm{\mu}_{\mathcal{V}} - \mathsf{R} 
 \right],
 \label{eq:KLD}
\end{align}
where the superscript ``KL'' is an abbreviation of ``Kullback-Leibler'', $\mbox{tr} \left\{ \cdot \right\}$ denotes the trace of a matrix, $\top$ denotes transposition, $\bm{0}_{\mathsf{R}}$ and $\bm{I}_{\mathsf{R}}$ are respectively the zero vector and unit matrix with dimension $\mathsf{R}$, and $\bm{\mu}_{\mathcal{V}}$ and $\bm{\Sigma}_{\mathcal{V}}$ are respectively the mean vector and covariance matrix calculated from $\bm{\mathrm{z}}$ of the various machine sounds. 
To generate anomalous sounds from (\ref{eq:prop_generator}), $\mathcal{G}$ needs to reconstruct various machine sounds, as $\bm{\mathrm{x}}^{(v)} = \mathcal{G} ( \mathcal{E} ( \bm{\mathrm{x}}^{(v)} \mid \Theta_{E} ) \mid \Theta_{G} )$. 
Thus, as a second constraint, we train $\Theta_E$ and $\Theta_G$ to minimize the reconstruction error (\ref{eq:mmse_ae}) calculated on the various machine sounds.

Next, we describe the GMM that models the PDF of the normal latent vectors. 
To reject a candidate $\tilde{\bm{\mathrm{z}}}$ which seems to be $\bm{\mathrm{z}}$ of a normal sound, we need to calculate the probability that the candidate is a normal one.
To calculate the probability, we need to model $p(\bm{\mathrm{z}} \mid y=0)$. 
Since there is no constraint on the form of $p(\bm{\mathrm{z}} \mid y=0)$ in the training procedure of $\Theta_E$, $p(\bm{\mathrm{z}} \mid y=0)$ might have a complex form. 
For simplicity, we use a GMM expressed as (\ref{eq:prp:log-likelihood}).

\subsection{Detailed description of training procedure}
\label{sec:full_procedure}

\begin{figure*}[ttt]
\centering
\includegraphics[width=175mm]{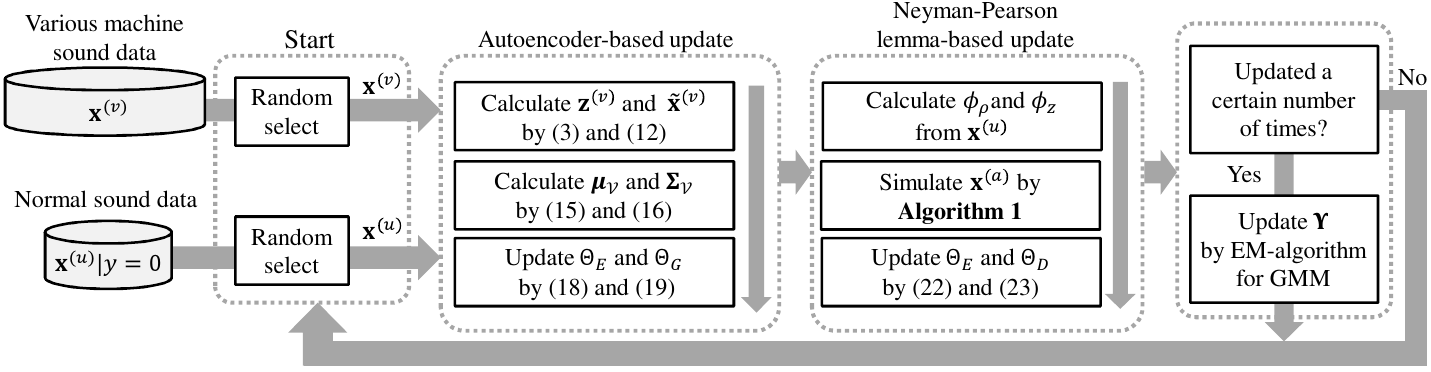}
\caption{Training procedure of the proposed method.}
\label{fig:train_procedure_np}
\end{figure*}

Here, we describe the details of the training procedure shown in Fig \ref{fig:train_procedure_np}. 
The training procedure consists in three steps. Hereafter, we call the proposed method using this training procedure {\tt NP-PROP}. 
The algorithm inputs are training data constructed from normal sounds and various machine sounds, and the outputs are $\Theta_E$ and $\Theta_D$. 
Moreover, $\bm{\mathrm{x}}^{(v)}_n$ and $\bm{\mathrm{x}}^{(u)}_n$ respectively denote the $n$-th training samples of minibatches of various and normal machine sounds, and $M$ is the number of samples included in a minibatch.

First, $\Theta_E$ and $\Theta_G$ are trained to simulate anomalous sounds. 
A minibatch of various machine sounds is randomly selected from the training dataset of various machine sounds. 
Next, its latent vectors are calculated as $\bm{\mathrm{z}}_{n}^{(v)} \gets \mathcal{E}(\bm{\mathrm{x}}_{n}^{(v)} | \Theta _E)$. 
Then, the parameters of the Gaussian distribution of the minibatch are calculated as
\begin{align}
\bm{\mu}_{\mathcal{V}} &= \frac{1}{M} \sum _{n=1}^{M} \bm{\mathrm{z}}_{n}^{(v)},\\
\bm{\Sigma}_{\mathcal{V}} &= \frac{1}{M} \sum _{n=1}^{M} 
\left( \bm{\mathrm{z}}_{n}^{(v)} - \bm{\mu}_{\mathcal{V}} \right)
\left( \bm{\mathrm{z}}_{n}^{(v)} - \bm{\mu}_{\mathcal{V}} \right)^{\top}.
\end{align}
Finally, to minimize the KLD and the reconstruction error of various sounds,
the objective function is calculated as 
\begin{align}
\mathcal{J}^{\mbox{\footnotesize KR}}(\Theta) 
= \mathcal{J}^{\mbox{\footnotesize KL}}(\Theta_{E}) + \sum_{n=1}^{M} 
\left\lVert \bm{\mathrm{x}}_{n}^{(v)} - \mathcal{G} \left( \mathcal{E} \left( \bm{\mathrm{x}}_{n}^{(v)} \mid \Theta _E \right) \mid \Theta _G \right) \right\rVert_2 ^2,
\label{eq:AE_update_prop}
\end{align}
where the superscript ``KR'' is an abbreviation of ``KLD and reconstruction'', 
and $\Theta_E$ and $\Theta_G$ are updated by gradient descent to minimize $\mathcal{J}^{\mbox{\footnotesize KR}}(\Theta) $:
\begin{align}
\Theta_E &\gets \Theta_E - \lambda \nabla_{\Theta_E} \mathcal{J}^{\mbox{\footnotesize KR}}(\Theta), \label{eq:propAE_update}\\
\Theta_G &\gets \Theta_G - \lambda \nabla_{\Theta_G} \mathcal{J}^{\mbox{\footnotesize KR}}(\Theta), 
\end{align}
where $\lambda$ is the step size.

Second, $\Theta_E$ and $\Theta_D$ are trained to maximize the objective function. 
A minibatch of normal sounds $\bm{\mathrm{x}}^{(u)}$ is randomly selected from the training dataset of normal sounds, and a minibatch of anomalous sounds $\bm{\mathrm{x}}^{(a)}$ is simulated using {\bf Algorithm \ref{alg:generate}}. 
Here, since DNN is not a parametric PDF, the threshold $\phi_{\rho}$ that satisfies $\mbox{FPR}(\Theta, \phi_{\rho}) = \rho$ cannot be analytically calculated. 
Thus, in this study, we approximately calculate $\phi_{\rho}$ by sorting the anomaly scores of normal sounds in the minibatch $\bm{\mathrm{x}}^{(u)}$. 
First, $\mathcal{A} (\bm{\mathrm{x}}^{(u)}, \Theta )$ and $-\ln (\bm{\mathrm{z}}^{(u)} \mid \Upsilon, y=0)$ are calculated, and $\phi_{\rho}$ and $\phi_z$ are set as the $\lfloor \rho M \rfloor$-th value of the sorted $\mathcal{A} (\bm{\mathrm{x}^{(u)}}, \Theta )$ and $-\ln (\bm{\mathrm{z}}^{(u)} \mid \Upsilon, y=0)$ in descending order, respectively. 
Then, the TPR and FPR are approximately evaluated as 
\begin{align}
\mbox{TPR}(\Theta, \phi_{\rho}) &\approx \frac{1}{M} \sum_{n=1}^{M} \mbox{sigmoid} \left(  \mathcal{A} \left( \bm{\mathrm{x}}^{(a)}_n, \Theta \right) - \phi_{\rho}  \right), \label{eq:TPR_ave}\\
\mbox{FPR}(\Theta, \phi_{\rho}) &\approx \frac{1}{M} \sum_{n=1}^{M} \mbox{sigmoid} \left(  \mathcal{A} \left( \bm{\mathrm{x}}^{(u)}_n, \Theta \right) - \phi_{\rho}  \right), \label{eq:FPR_ave}
\end{align}
where the binary decision function $\mathcal{H}$ is approximated by a sigmoid function, allowing the gradient to be analytically calculated.
Finally, $\Theta_E$ and $\Theta _D$ are updated to increase $\mathcal{J}^{\mbox{\footnotesize NP}}(\Theta)$ by gradient ascent:
\begin{align}
\Theta_E &\gets \Theta_E + \lambda \nabla_{\Theta_E} \mathcal{J}^{\mbox{\footnotesize NP}}(\Theta), \\
\Theta_D &\gets \Theta_D + \lambda \nabla_{\Theta_D} \mathcal{J}^{\mbox{\footnotesize NP}}(\Theta). \label{eq:propNP_update}
\end{align}

Third, to update the PDF of the latent vectors of normal sounds $p(\bm{\mathrm{z}} \mid \Upsilon, y=0)$, when (\ref{eq:propAE_update})--(\ref{eq:propNP_update}) is repeated a certain number of times, $\Upsilon$ is updated using the expectation-maximization (EM) algorithm for GMM using all training data of normal sounds. 
The above algorithm is run a pre-defined number of epochs.

\subsection{Detailed description of detection procedure}
\label{sec:full_procedure_detect}

After training $\Theta_E$ and $\Theta_D$, we can identify whether the observed sound is a normal one or not. 
First, the input vector $\bm{\mathrm{x}}_{\tau}, \tau \in \{ 1,..., T\}$ is calculated from the observed sound. 
Then, the anomaly score is calculated as (\ref{eq:anm_score_AE}). Finally, a decision score, 
$
V = \frac{1}{T} \sum_{\tau = 1 }^{T} \mathcal{H}(\bm{\mathrm{x}}_{\tau}, \Theta, \phi ),
$
is calculated, and when $V$ exceeds a pre-defined value $\phi_V$, the observed sound is determined to be anomalous. 
In this study, we used $\phi_V = 0$, meaning that, if the anomaly score exceeds the threshold even for one frame, the observed sound is determined to be anomalous.

\subsection{Modified implementation as an AUC maximization}
\label{sec:NPextension}
The receiver operating characteristic (ROC) curve and the AUC are widely used performance measures for imbalanced data classification and/or anomaly detection. 
The AUC is calculated as 
\begin{align}
\mbox{AUC}(\Theta) &= 
\mathbb{E} \left[ 
\underbrace{
\mathbb{E} \left[
\mathcal{H}(\bm{\mathrm{x}}', \Theta , \mathcal{A}(\bm{\mathrm{x}}, \Theta))
\right]_{ \bm{\mathrm{x}}'|y \neq 0 }
}_{
\mbox{TPR}
\left(
\Theta, \mathcal{A}
\left( \bm{\mathrm{x}}, \Theta \right)
\right)
}
\right]_{ \bm{\mathrm{x}} |y=0 }
, \\
&\approx \frac{1}{M} \sum_{n=1}^{M} 
\mbox{TPR}
\left(
\Theta, \mathcal{A}
\left( \bm{\mathrm{x}}_n^{(u)} , \Theta \right)
\right).
\label{eq:AUC_ave}
\end{align}
As we can see in (\ref{eq:AUC_ave}), anomalous sound data are needed to calculate the AUC. 
Although the AUC has been used as an objective function in imbalanced data classification \cite{Bradley_1997,Herschtal_2004,Fujino_2016}, it has not been applied to unsupervised-ADS so far. 
Fortunately, since the proposed rejection sampling can simulate anomalous sound data, AUC maximization can be used as an objective function of ADS. 
Instead of $\mathcal{J}^{\mbox{\footnotesize NP}}(\Theta)$, the following objective function can be used in the training procedure:
\begin{align}
\begin{split}
&\mathcal{J}^{\mbox{\footnotesize AUC}}(\Theta) \\
&= \frac{1}{M} \sum_{n=1}^{M} 
\mbox{TPR} \left( \Theta, \mathcal{A}
\left( \bm{\mathrm{x}}_n^{(u)} , \Theta \right)
 \right)
-
\mbox{FPR} \left( \Theta, \mathcal{A}
\left( \bm{\mathrm{x}}_n^{(u)} , \Theta \right)
\right).
\end{split}
\label{eq:AUC_prop}
\end{align}
Hereafter, we call the proposed method using $\mathcal{J}^{\mbox{\footnotesize AUC}}(\Theta) $ instead of $\mathcal{J}^{\mbox{\footnotesize NP}}(\Theta)$ {\tt AUC-PROP}.

\section{Experiments}
\label{sec:eval}

We conducted experiments to evaluate the performance of the proposed method. 
First, we conducted an objective experiment using synthetic anomalous sounds (Sec. \ref{sec:exp_data}). 
To generate a large enough anomalous dataset for the ADS accuracy evaluation, we used collision and sustained sounds from datasets for {\it detection and classification of acoustic scenes and events 2016} (DCASE-2016 \cite{DCASE2016}). 
To show the effectiveness of the method in real environments, we conducted verification experiments in three real environments (Sec. \ref{sec:verifi}).

\subsection{Experimental conditions}
\label{sec:exp_cond}

\subsubsection{Compared methods}

The proposed methods described in Sec \ref{sec:full_procedure} ({\tt NP-PROP}) and Sec \ref{sec:NPextension} ({\tt AUC-PROP}) were compared with three state-of-the-art ADS methods: 
\begin{itemize}
\item {\tt AE} \cite{Marchi_2015}: ADS using the autoencoder described in Sec \ref{sec:AE_ADS}. The encoder and decoder were trained to minimize (\ref{eq:mmse_ae_sum}).
\item {\tt VAE} \cite{VAE_Anomaly}: $\mathcal{E}$ and $\mathcal{D}$ were implemented using VAE. 
The encoder estimated the mean and variance parameters of the Gaussian distribution in the latent space. 
Then, the latent vectors were sampled from the Gaussian distribution whose parameters were estimated by the encoder. 
Then, the decoder reconstructed the input vector from the sampled latent vectors. Finally, the reconstruction error was calculated and used as the anomaly score.
\item {\tt VAEGAN} \cite{VAEGAN}: To investigate the effectiveness of the anomalous sound simulation, VAEGAN \cite{VAEGAN} was used to simulate fake normal data. 
The generators ({\it i.e.} VAE) were used to simulate fake normal sounds. The output of the discriminator without the sigmoid activation was used as the anomaly score.
\end{itemize}
We also used our previous work \cite{Koizumi_2017_ADS} ({\tt CONV-PROP}) for comparison. 
This method uses a VAE to extract latent vectors as acoustic features. 
A GMM is used for the normal model, and the encoder and decoder are trained to maximize (\ref{eq:NP_obj_proposed}).

\vspace{5pt}
\subsubsection{DNN architecture and setup}
\begin{figure}[ttt]
  \centering
  \includegraphics[width=90mm]{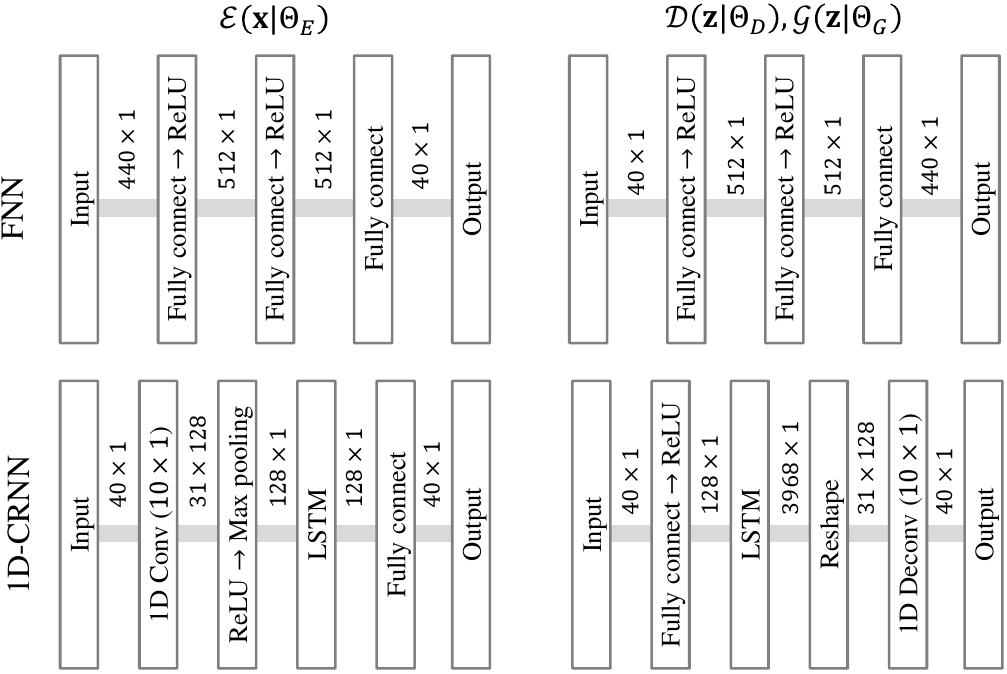}
  \caption{
Network architectures of encoder, decoder and generator used for {\tt NP-PROP}, and {\tt AUC-PROP}. 
The encoder and decoder of {\tt AE} have the same architecture.  
In {\tt VAE}, {\tt VAEGAN} and {\tt CONV-PROP}, the encoder has two output layers for the mean and variance vector.
In {\tt VAEGAN}, the architecture of the discriminator is the same as that of the encoder, but the output dimension of the fully connected layer is 1.
}
  \label{fig:architecture}
\end{figure}

We tested two types of network architecture as shown in Fig. \ref{fig:architecture}. 
The first architecture, ``FNN'', consisted of fully connected DNNs with three hidden layers and 512 hidden units. 
The rectified linear unit (ReLU) was used as the activation functions of the hidden layers. 
The input vector $\bm{\mathrm{x}}$ was defined as
\begin{align}
\nonumber
\bm{\mathrm{x}}_{\tau} &:= \left( 
\ln \left[ \mbox{Mel} \left[\mbox{Abs} \left[ \bm{\mathrm{X}}_{\tau - C} \right] \right] \right]
, ...,
\ln \left[ \mbox{Mel} \left[\mbox{Abs} \left[ \bm{\mathrm{X}}_{\tau + C} \right] \right] \right]
\right)^{\top},\\
\nonumber
\bm{\mathrm{X}}_{\tau} &:= \left(
X_{1,\tau}, ..., X_{\Omega, \tau}
 \right),
\end{align}
where $X_{\omega, \tau}$ is the discrete Fourier transform (DFT) spectrum of the observed sound, $\omega \in \{ 1,...,\Omega \}$ denotes the frequency index, $C (= 5$) is the context window size, and $\mbox{Mel}[\cdot]$ and $\mbox{Abs}[\cdot]$ denote 40-dimensional Mel matrix multiplication and the element-wise absolute value. 
Thus, the dimension of $\bm{\mathrm{x}}$ was $\mathsf{Q} = 40 \times (2C+1) = 440$. 
The second architecture,``1D-CRNN'', consisted in a one-dimensional convolution neural network (1D-CNN) layer and a long short-term memory (LSTM) layer; it worked well in supervised anomaly detection (race SED) in DCASE 2017 \cite{rare_01}. In order to detect anomalous sounds in real time, we changed the backward LSTM to a forward one. 
In addition, to avoid overfitting, we used only one forward LSTM layer instead of two backward LSTM layers. 
The input vector $\bm{\mathrm{x}}$ was a 40-dimensional log mel-band energy:
\begin{align}
\nonumber
\bm{\mathrm{x}}_{\tau} := \ln \left( 
\mbox{Mel} \left[\mbox{Abs} \left[ \bm{\mathrm{X}}_{\tau} \right] \right] 
\right)^{\top}.
\end{align}
The dimension of $\bm{\mathrm{x}}$ was $\mathsf{Q} = 40$. 
For each architecture, the dimension of the latent vector $\bm{\mathrm{z}}$ was $\mathsf{R} = 40$. 
All input vectors were mean-and-variance normalized using the training data statistics.

As an implementation for the gradient method, the Adam method \cite{adam} was used instead of the gradient descent/ascent shown in (\ref{eq:propAE_update})--(\ref{eq:propNP_update}). 
To avoid overfitting, $L_2$ normalization \cite{l2_devay} with a regularization penalty of $10^{-4}$ was used. 
The minibatch size for all methods was $M = 512$. 
All models were trained for 500 epochs.
In all methods, the average value of the loss was calculated on the training set at every epoch, and when the loss did not decrease for five consecutive epochs, the stepsize was decreased by half.

\vspace{5pt}
\subsubsection{Other conditions}

All sounds were recorded at a sampling rate of 16 kHz. The frame size of the DFT was 512, and the frame was shifted every 256 samples. 
For $p(\bm{\mathrm{z}} \mid \Upsilon, y=0)$, the number of Gaussian mixtures was $K = 16$ and a diagonal covariance matrix was used to prevent the problem from being ill-conditioned. 
The EM algorithm for the GMM involved iterating (\ref{eq:propAE_update})--(\ref{eq:propNP_update}) 30 times. 
All the above-mentioned conditions are summarized in Table \ref{tbl:param_tbl}. 

\begin{table}[tt]
\centering
\caption{Experimental conditions}
\small
\begin{tabular}{l|l}\hline \hline
	\multicolumn{2}{c}{Parameters for signal processing}\\ \hline 
 Sampling rate							& 16.0 kHz \\
 FFT length			 					& 512 pts	\\
 FFT shift length	 					& 256 pts	\\ 
 Number of mel-filterbanks 		& 40 		\\ \hline 
 \multicolumn{2}{c}{Other parameters}\\ \hline 
 Context window size $C$											& 5 \\ 
 Dimension of input vector $\mathsf{Q}$ for FNN 	& 440 \\
 Dimension of input vector $\mathsf{Q}$ for 1D-CRNN 	& 40 \\ 
 Dimension of acoustic feature vector $\mathsf{R}$ 	& 40 \\ 
 GMM update per gradient method			& $30$ \\ 
 Number of mixtures $K$ 							& 16 \\ 
 Minibatch size $M$ 							& 512\\
 FPR parameter $\rho$						& 0.2\\
 Step size $\lambda$								& $10^{-4}$	\\ 
 $L_2$ normalization parameter				& $10^{-4}$ \\ 
\hline 
\hline
\end{tabular}
\normalsize
\label{tbl:param_tbl}
\end{table}

\subsection{Objective experiments on synthetic data}
\label{sec:exp_data}

\subsubsection{Dataset}

\begin{figure*}[ttthhh]
  \centering
  \includegraphics[width=140mm]{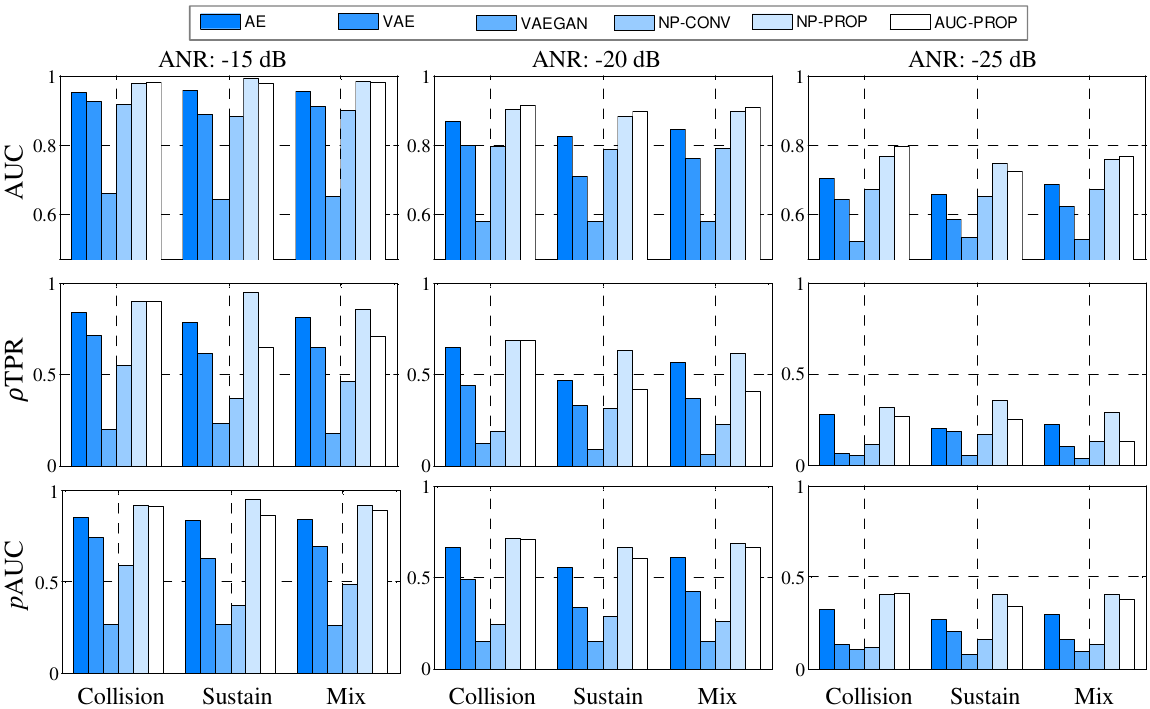}
  \caption{
Evaluation results of FNN. 
}
  \label{fig:fnn_obj}
  \includegraphics[width=140mm]{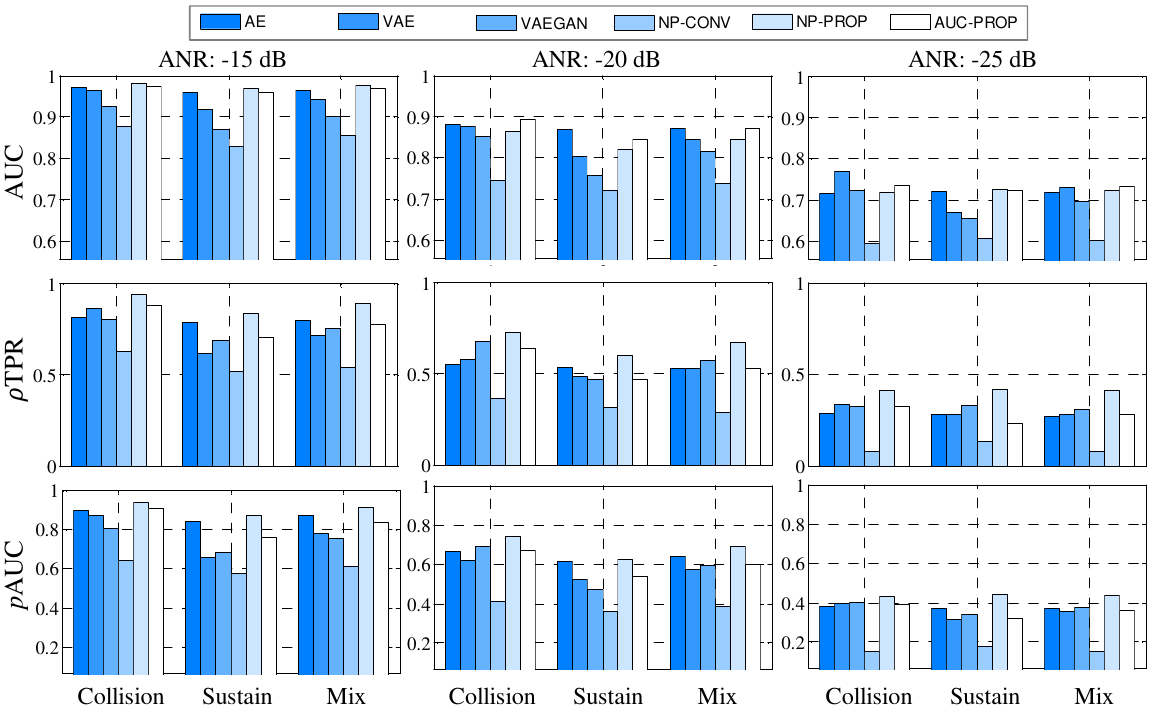}
  \caption{
Evaluation results of 1D-CRNN. 
}
  \label{fig:crnn_obj}
\end{figure*}
\begin{figure*}[ttthhh]
  \centering
  \includegraphics[width=145mm]{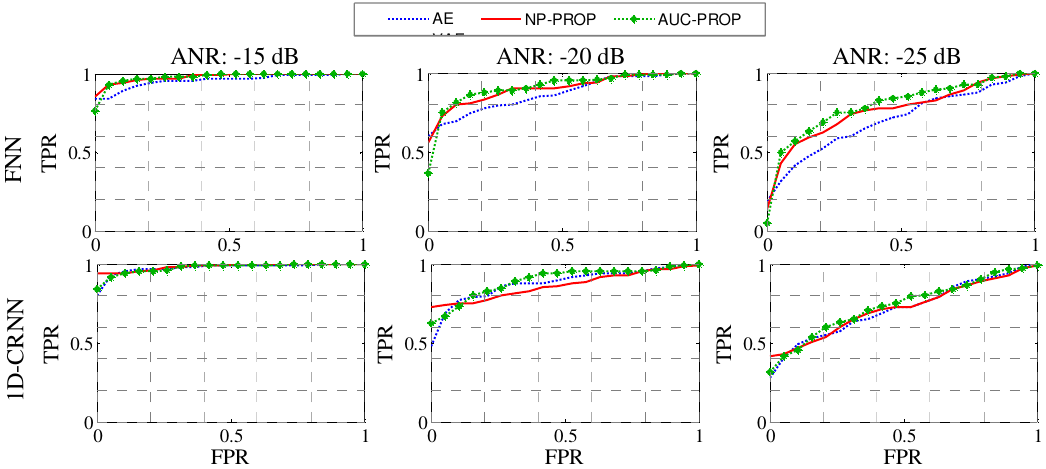}
  \caption{ROC curves of {\tt AE}, {\tt NP-PROP} and {\tt AUC-PROP} for each ANR condition evaluated on Mix dataset.}
  \label{fig:all_roc}
\end{figure*}

Sounds emitted from a condensing unit of an air conditioner operating in a real environment were used as the normal sounds. 
In addition, various machine sounds were recorded from other machines, including a {\it compressor}, {\it engine}, {\it compression pump}, and an {\it electric drill}, as well as environmental noise of factories.
The normal and various machine sound data totaled 4 and 20 hours ($=$ 4 hours normal $+$ 16 hours other machines), respectively. 
These sounds were recorded at a 16-kHz sampling rate. 
In order to improve the robustness for different loudness levels and ratios of the normal and anomalous sound, the various machine sounds in the training dataset were augmented with a multiplication of five amplitude gains.
These gains are calculated so that the maximum amplitudes of various sounds becomes to 1.0, 0.5, 0.25, 0.125, and 0.063.

Since it is difficult to collect a massive amount of test data including anomalous sounds, synthetic anomalous data were used in this evaluation. 
In particular, we used the training datasets for task of DCASE-2016 \cite{DCASE2016} as anomalous sounds. 
Although these sounds are ``normal'' sounds in an office, in unsupervised-ADS, the unknown  sounds are categorized as ``anomalous''. 
Thus, we consider that this evaluation can at least evaluate the detection performance for unknown sounds. 
Since the anomalous sounds of machines are roughly categorized into collision sounds ({\it e.g.}, the sound of a metal part falling on the floor) and sustained sounds ({\it e.g.}, frictional sound caused by scratched bearings), we selected 80 collision sounds, including ({\it slamming doors }, {\it knocking at doors }, {\it keys put on a table}, {\it keystrokes on a keyboard}), and 60 sustained sounds ({\it drawers being opened}, {\it pages being turned}, and {\it phones ringing}), from this dataset \cite{dl_url}. 
To synthesize the test data, the anomalous sounds were mixed with normal sounds at anomaly-to-normal power ratios (ANRs\footnote{ ANR is a measure comparing the level of an anomalous sound to the level of a normal sound. This definition is the same as the signal-to-noise ratio (SNR) when the signal is an anomalous sound and the noise is a normal sound.}) of -15, -20 and -25 dB using the following procedure:
\begin{enumerate}
\item select an anomalous sound and randomly cut a normal so that has the same signal length of the selected anomalous sound.
\item for the cut normal and anomalous sounds, calculate the frame-wise log power of each of 512 points with a 256 point shift on a dB scale, namely $ \mathcal{P}_{\tau} = 20 \log _{10} \sum _{\omega = 1}^{\Omega} \left| X_{\omega, \tau} \right|. $
\item select the median of $\mathcal{P}_{\tau}$ as the representative power of each sound as.
\item manipulate the power of the anomalous sound so that the ANR has the desired value.
\item used the cut  normal sound as the test data of normal sound, and generate the test data of the anomalous sound by mixing the anomalous sound with the quarried normal sound.
\end{enumerate}
In total, we used 140 normal and anomalous sound samples for each ANR condition. 
The training dataset of normal sounds and the MATLAB code to generate the test dataset are freely available on the website\footnote{\url{https://archive.org/details/ADSdataset}}.

\begin{figure*}[ttt]
  \centering
  \includegraphics[width=180mm,clip]{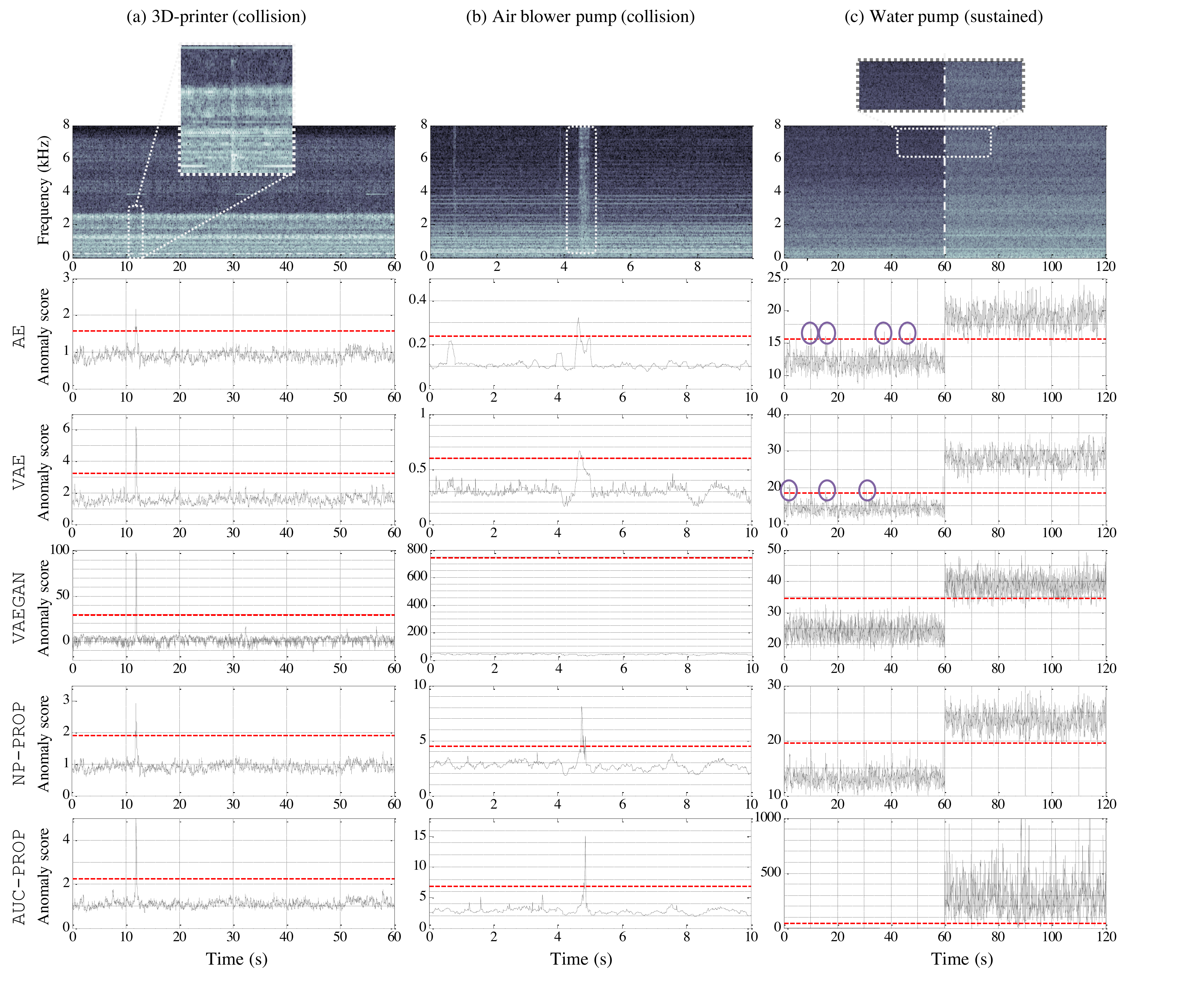}  
  \caption{
Anomaly detection results for sound emitted from 3D-printer (left), air blower pump (center), and water pump (right). 
The top figure shows the spectrogram, and the bottom figures show the anomaly score (black solid line) and threshold $\phi_{0.001}$ (red dashed line) of each method. 
Anomalous sounds are enclosed in white dotted boxes, and false-positive detections are circled in purple. 
Since the spectrum changes due to the anomalous sounds of 3D-printer and water pump are difficult to see, their anomalous sounds are enlarged. 
In addition, since anomalous sound of the water pump is a sustained, 60 seconds of normal sounds and 60 seconds of anomalous sound are concatenated for comparison. 
}
  \label{fig:real_result}
\end{figure*}

\vspace{5pt}
\subsubsection{Results}

To evaluate the performance of ADS, we used the AUC, $\rho$TPR, and partial AUC ($p$AUC) \cite{pAUC}. 
The AUC is a traditional performance measure of anomaly detection.
The other two measurements evaluated the performance under low FPR conditions. 
$\rho$TPR is the TPR under the condition that FPR equals $\rho$. The $p$AUC is an AUC calculated with FPRs ranging from 0 to $p$ with respect to the maximum value of $1$. 
The parameters were $\rho = 0.05$ and $p = 0.1$. 
We evaluated these metrics for three different evaluation sets: 80 collision sounds (Collision), 60 sustained sounds (Sustain), and the sum of these $80+60=140$ sounds (Mix).

The results for each score, sound category, and ANR on FNN and 1D-CRNN are shown in Fig. \ref{fig:fnn_obj} and Fig. \ref{fig:crnn_obj}. 
Overall, the performances of {\tt AE}, {\tt NP-PROP} and {\tt AUC-PROP} were better than those of {\tt VAE} and {\tt VAEGAN}. 
In detail, {\tt AE} achieved high scores for all measurements, {\tt AUC-PROP} achieved high scores for AUC and $p$AUC, and {\tt NP-PROP} achieved high scores for $\rho$TPR and $p$AUC. 
In addition, for all conditions, the $\rho$TPR and $p$AUC scores of {\tt NP-PROP} were higher than those of {\tt AE}. 
To discuss the difference between the objective functions of {\tt AE}, {\tt NP-PROP} and {\tt AUC-PROP}, we show the ROC curves in Fig. \ref{fig:all_roc}. 
Since the differences between the results of Collision, Sustained, and Mix were small, we plotted only those of the Mix dataset. 
From these ROC curves, we can see that the TPRs of {\tt NP-PROP} under the low FPR conditions were significantly higher than those of other methods. 
This might be because the objective function of {\tt NP-PROP} works to increase TPR under the low FPR condition. 
In addition, although {\tt AUC-PROP}'s TPRs under the low FPR condition were lower than those of {\tt NP-PROP}, the TPRs under the moderate and high FPR conditions were higher than those of the other methods. 
This might be because the objective function of {\tt AUC-PROP} works to increase TPR for all FPR conditions. 
Since the individual results and objective function tend to coincide, we consider that the training of each neural network succeeded. 
In addition, TPR under the low FPR conditions is especially important when the ADS is used in real environments, because if an ADS system frequently gives false alert, we cannot trust it. 
Therefore, unsupervised-ADS using an AE trained using (\ref{eq:NP_obj_proposed}) would be effective in real situations.

In addition, regarding the FNN results, {\tt VAE} scored lower than {\tt AE}, and {\tt VAEGAN} scored lower than all the other methods. 
These results suggest that when calculating the anomaly score using a simple network architecture like FNN, a simple reconstruction error would be better than complex calculation procedures such as {\tt VAE} and {\tt VAEGAN}. 
Moreover, the scores of {\tt NP-CONV} were lower than those of the DNN-based methods. 
In our previous study \cite{Koizumi_2017_ADS}, we used a DNN a feature extractor and constructed the normal model by using a GMM. 
These results suggest that using a DNN for the normal model would be better than using a GMM.

\subsection{Verification experiment in a real environment}
\label{sec:verifi}

We conducted three verification experiments to test whether anomalous sounds in real environments can be detected. 
The target equipment and experimental conditions were as follows:
\begin{itemize}
\item Stereolithography 3D-printer: 
We collected an actual collision-type anomalous sound. 
Two hours worth of normal sounds were collected as training data. 
The anomalous sound was caused by collision of the sweeper and the formed object.
The 3D-printer stopped 5 minutes after this anomalous sound occurred.
\item Air blower pump: We collected an actual collision-type anomalous sound. 
Twenty minutes worth of normal sounds were collected as training data. 
The anomalous sound was caused by blockage by a foreign object stuck in the air blower duct. 
This anomaly does not lead to immediate machine failure; however, it should be addressed.
\item Water pump: We collected an actual sustained type anomalous sound. 
Three hours worth of normal sounds were collected as training data. 
Above 4 kHz, the anomalous sound has a larger amplitude than that of the normal sounds, and it was due to wearing of the bearings. 
An expert conducting a periodic inspection diagnosed that the bearings needed to be replaced.
\end{itemize}
All anomalous and normal sounds were recorded at a 16-kHz sampling rate. The other conditions were the same as in the objective experiment. The FNN architecture was used for the anomaly score calculation.

Figure \ref{fig:real_result} shows the spectrogram (top) and anomaly scores of each method (bottom). 
The red dashed line in each of the bottom figures is the threshold $\phi_{0.001}$, which is defined such that the FPR of the training data was 0.1\%. 
Anomalous sounds are enclosed in white dotted boxes in the spectrograms, and the false-positive detections are circled in purple in the anomaly score graphs. 
Since the anomalous sound of the water pump is a sustained sound, for ease of comparison, 60 seconds of normal sounds and 60 seconds of anomalous sound are concatenated in each figure. 
In addition, the anomalous sounds are enlarged, since the spectrum changes due to the anomalous sounds of the 3D-printer and water pump are difficult to see.

All of the results for {\tt NP-PROP} and {\tt AUC-PROP} indicate that anomalous sounds were clearly detected; the anomaly scores of the anomalous sounds evidently exceeded the threshold, while those of the normal sounds were below the threshold. 
Meanwhile, in the results of {\tt AE} and {\tt VAE}, although the anomaly scores of all anomalous sounds exceeded the threshold, false-positives were also observed in the results for the water pump. 
In addition, although {\tt AE}'s anomaly score of the 3D-printer and {\tt VAE}'s anomaly score of the air blower pump exceeded the threshold, the excess margin of the anomaly score is small and it is difficult to use a higher threshold for reducing FPR. 
This problem  might be because that the objective functions do not work to increase anomaly scores for anomalous sounds, and thus, the encoder and decoder reconstructed not only normal sounds but also anomalous sounds. 
In {\tt VAEGAN}, the anomaly scores of the 3D-printer and the water pump exceeded the threshold, whereas those of the air blower pump did not exceed the threshold. 
The reason might be that when the generator precisely generates ``fake'' normal sounds, the normal model is trained to increase the anomaly scores of normal sounds. 
Therefore, the threshold of the air blower pump, which is defined as the FPR of normal training data becoming 0.001, takes a very high value.  
These verification experiments suggest that the proposed method is effective at identifying anomalous sounds under practical conditions.

\section{Conclusions}
\label{sec:conclusion}
This paper proposed a novel training method for unsupervised-ADS using an AE for detecting unknown anomalous sound. 
The contributions of this research are as follows: 
1) by considering outlier-detection-based ADS as a statistical hypothesis test, we defined an objective function that builds upon the Neyman-Pearson lemma \cite{Ney_Pear}. 
The objective function increases the TPR under a low FPR condition, which is often used in practice. 
2) By considering the set of anomalous sounds to be complement to the set of normal sounds, we formulated a rejection sampling algorithm to simulate anomalous sounds. 
Experimental results showed that these contributions enabled us to construct an ADS system that accurately detects unknown anomalous sounds in three real environments. 

In future, we will tackle the following remaining issues of ADS systems in real environments:

1) Extension to a supervised approach to detect both known and unknown anomalous sounds: 
while operating an ADS system in a real environment, we may occasionally obtain partial samples of anomalous sounds. 
While it might be better to use the collected anomalous sounds in training, the cross-entropy loss would not be the best way to detect both known and unknown anomalous sounds \cite{Gornitz2013}. 
In addition, if we calculate the TPR in $\mathcal{J}^{\mbox{\footnotesize NP}}(\Theta)$ and/or $\mathcal{J}^{\mbox{\footnotesize AUC}}(\Theta)$ only using a part of the anomalous sounds, this training does not guarantee the performance for unknown anomalous sounds.
Thus, we should develop a supervised-ADS method that can also detect unknown anomalous sounds; a preliminary study on this has been published in \cite{Kawachi_2018}.

2) Incorporating machine or context-specific knowledge: to simplify the experiments, we used the simple detection rule described in Sec. \ref{sec:full_procedure_detect}. 
However, for the anomaly alert, it would be better to use machine/context-specific rules, such as modifying or smoothing the detection result from the raw anomaly score. 
Thus, it will be necessary to develop rules or a trainable post-processing block to modify the anomaly score.


\newpage
\appendix
\subsection{List of Symbols}
\label{sec:app_A}

1. Functions

\vspace{5pt}

\begin{tabular}{cp{0.6\textwidth}}
  $\mathcal{J}$ 				& Objective function \\
  $\mathcal{A}$ 				& Anomaly score \\
  $\mathcal{H}$ 				& Binary decision \\
  $\mathcal{E}$ 				& Encoder of autoencoder \\
  $\mathcal{D}$ 				& Decoder of autoencoder \\
  $\mathcal{G}$ 				& Generator \\
  $\mathcal{N}$ 			& Gaussian distribution \\
  $\mathbb{E} [\cdot]_x$ 	& Expectation with respect to $x$ \\
  $\nabla_{x} (\cdot)$ 		& Gradient with respect to $x$ \\
  $\mbox{tr} (\cdot)$ 		& Trace of matrix \\
  $D(A || B)$ 						& Kullback-Leibler divergence between $A$ and $B$\\
  $\lVert \cdot \rVert _2$ 		& $L_2$ norm\\
  $\lfloor \cdot \rfloor$ 		& Flooring function\\
\end{tabular}\\

2. Parameters

\vspace{5pt}

\begin{tabular}{cp{0.6\textwidth}}
  $\Theta$ 		& Parameters of normal model \\
  $\Theta_E$ 	& Parameters of encoder \\
  $\Theta_D$ 	& Parameters of decoder \\
  $\Theta_G$ 	& Parameters of generator \\
  $\Upsilon$ 	& Parameters of Gaussian mixture model \\
\end{tabular}\\

3 Variables

\vspace{5pt}

\begin{tabular}{cp{0.6\textwidth}}
 $\bm{\mathrm{x}}$ 	& Input vector \\
 $y$ 								& State variable \\
 $\bm{\mathrm{z}}$ 	& Latent vector \\
 $\phi$ 							& Threshold for anomaly score \\
 $\rho$ 							& Desired false positive rate \\
 $\bm{\mu}$ 		& Mean vector \\
 $\bm{\Sigma}$ 	& Covariance matrix \\
 $w$ 						& Mixing weight of Gaussian mixure model\\
 $K$ 						& Number of gaussian mixtures \\
 $T$ 								& Number of time frames of observation \\
 $N$ 								& Number of training samples \\
 $M$ 								& Minibatch size \\
 $\mathsf{Q}$ 	& Dimension of input vector \\
 $\mathsf{R}$ 	& Dimension of latent vector \\
 $\lambda$ 				& Step size for gradient method \\
 $C$ 							& Context window size \\
 $\ell$ 							& Temporary variable of anomaly score\\
 $V$ 							& Anomaly decision score for one audio clip \\
\end{tabular}\\

4. Notations

\vspace{5pt}

\begin{tabular}{cp{0.6\textwidth}}
 $\tau$ 	& Time-frame index of observation\\
 $n$ 		& Index of training sample\\
 $k$ 		& Index of Gaussian distribution\\
 $(\cdot)^{\top}$ 	& Transpose of matrix or vector \\
 $(\cdot)^{(u)}$ 	& Variable of normal sound \\
 $(\cdot)^{(a)}$ 	& Variable of anomalous sound \\
 $(\cdot)^{(v)}$ 	& Variable of various sound \\
\end{tabular}

\begin{IEEEbiography}
[{\includegraphics[width=1in,height=1.25in,clip,keepaspectratio]{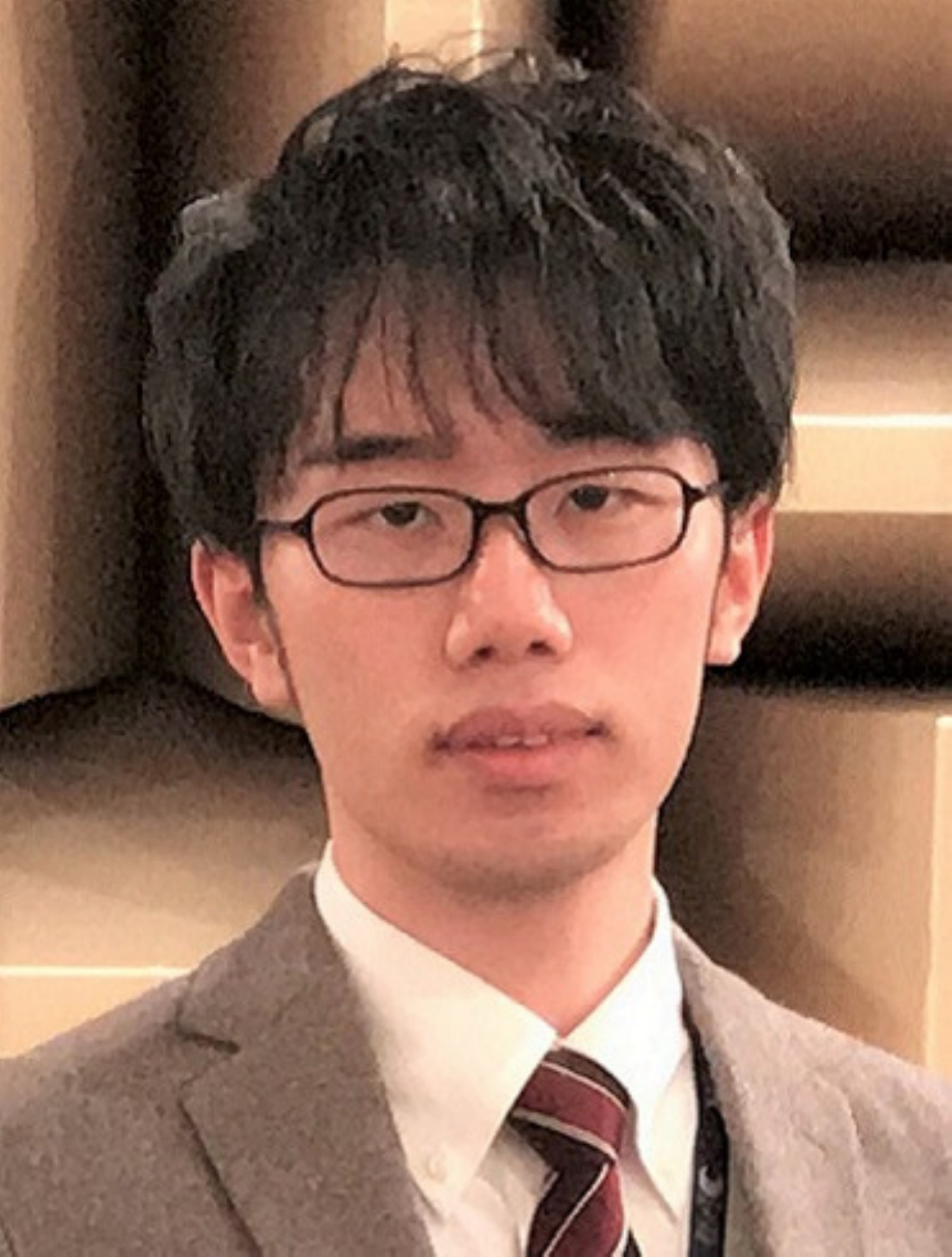}}]
{Yuma Koizumi}(M '15) received the B.S. and M.S. degrees from Hosei University, Tokyo, in 2012 and 2014, and the Ph.D. degree from the University of Electro-Communications, Tokyo, in 2017.
Since joining the Nippon Telegraph and Telephone Corporation (NTT) in 2014, 
he has been researching acoustic signal processing and machine learning including basic research of sound source enhancement and unsupervised/supervised anomaly detection in sounds.
He was awarded the FUNAI Best Paper Award and the IPSJ Yamashita SIG Research Award from the Information Processing Society of Japan (IPSJ) in 2013 and 2014, respectively, and the Awaya Prize from the Acoustical Society of Japan (ASJ) in 2017. 
He is a member of the ASJ and the Institute of Electronics, Information and Communication Engineers (IEICE).
\end{IEEEbiography}
\begin{IEEEbiography}
[{\includegraphics[width=1in,height=1.25in,clip,keepaspectratio]{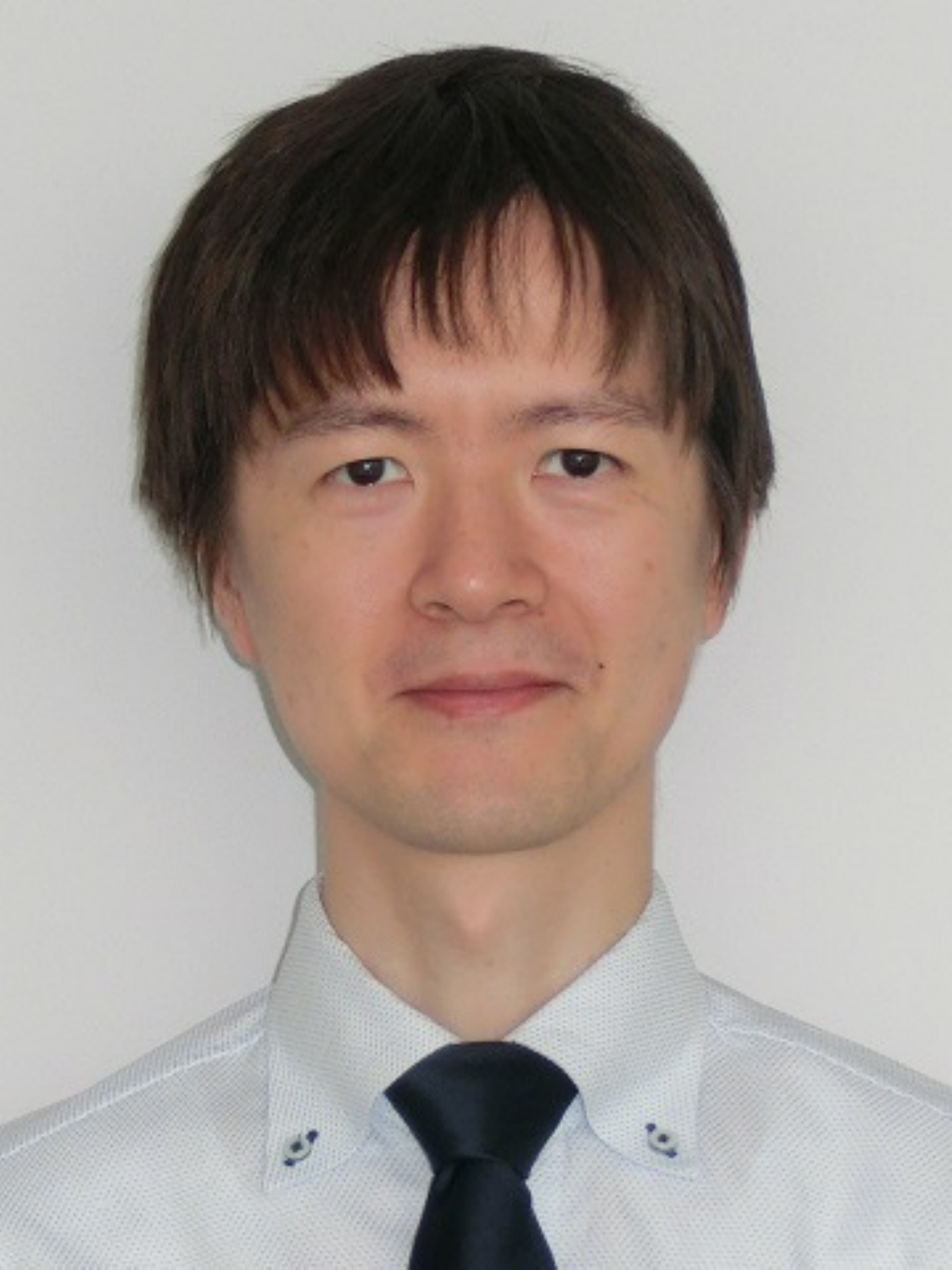}}]
{Shoichiro Saito} (SM '06-M '07) received the B.E. and M.E. degrees from the University of Tokyo in 2005 and 2007.
Since joining NTT in 2007, he has been engaging in research and development of acoustic signal processing systems including acoustic echo cancellers, hands-free telecommunication, and anomaly detection in sound. 
He is currently a Senior Research Engineer of Audio, Speech, and Language Media Laboratory, NTT Media Intelligence Laboratories. 
He is a member of the IEICE, and the ASJ.
\end{IEEEbiography}
\begin{IEEEbiography}
[{\includegraphics[width=1in,height=1.25in,clip,keepaspectratio]{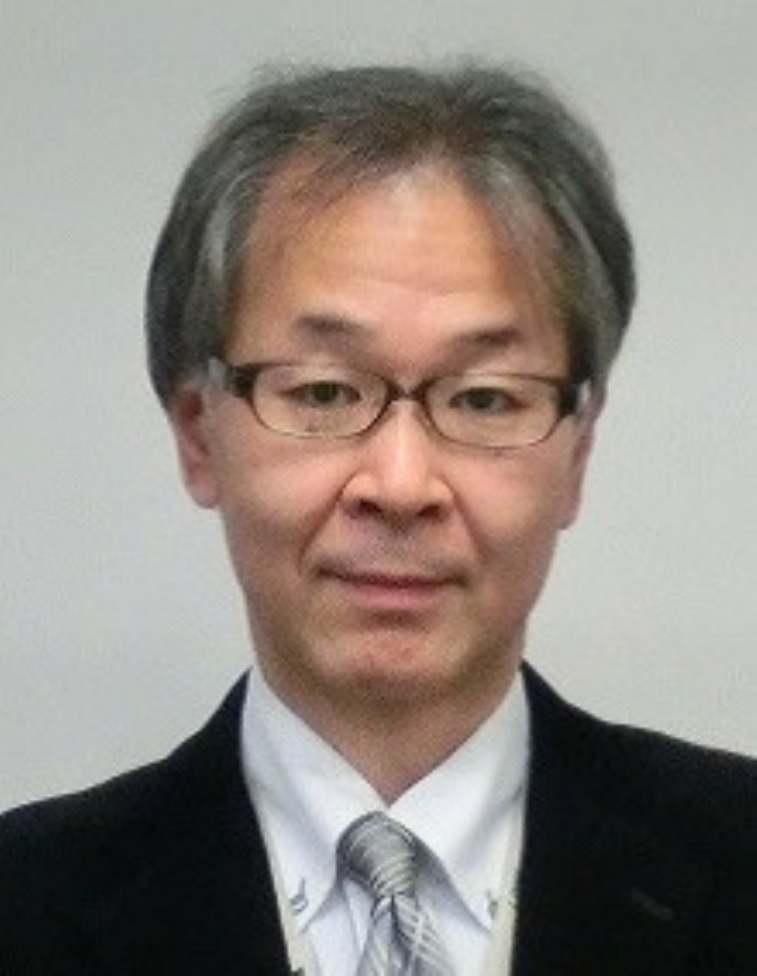}}]
{Hisashi Uematsu} received the B.E., M.E., and Ph.D. degrees in Information Science from Tohoku University, Miyagi, in 1991, 1993, and 1996. 
He joined NTT in 1996 and has been engaged in research on psycho-acoustics (human auditory mechanisms) and digital signal processing.  
He is currently a Senior Research Engineer of Cross-Modal Computing Project, NTT Media Intelligence Laboratories.
He was awarded the Awaya Prize from the ASJ in 2001. He is a member of the ASJ. 
\end{IEEEbiography}
\begin{IEEEbiography}
[{\includegraphics[width=1in,height=1.25in,clip,keepaspectratio]{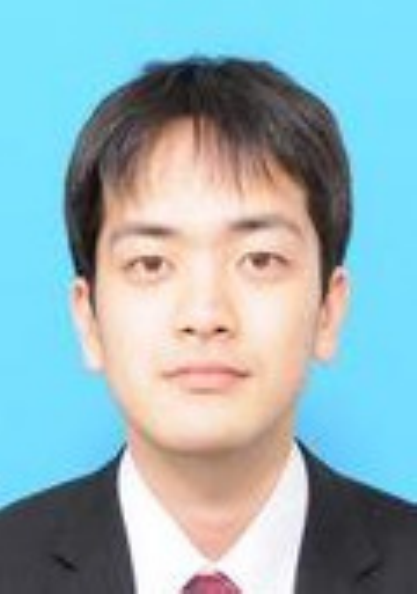}}]
{Yuta Kawachi} received a B.E. and M.E. degrees from Waseda University, Tokyo, in 2012 and 2014. 
Since joining NTT in 2014, he has been researching acoustic signal processing and machine learning. 
He is a member of the ASJ. 
\end{IEEEbiography}
\begin{IEEEbiography}
[{\includegraphics[width=1in,height=1.25in,clip,keepaspectratio]{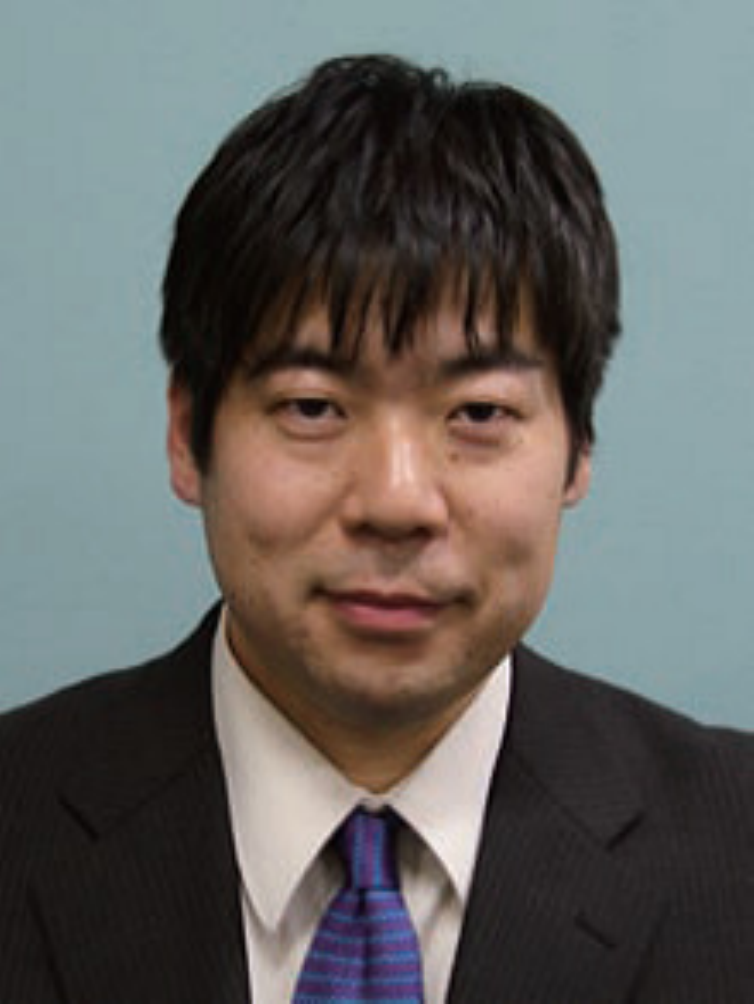}}]
{Noboru Harada} (M '99-SM '18) received the B.S., and M.S., degrees from the Department of Computer Science and Systems Engineering of Kyushu Institute of Technology in 1995 and 1997, respectively. 
He received the Ph.D. degree from the Graduate School of Systems and Information Engineering, University of Tsukuba in 2017. 
Since joining NTT in 1997, he has been researching speech and audio signal processing such as high efficiency coding and lossless compression. 
His current research interests include acoustic signal processing and machine learning for acoustic event detection including anomaly detection in sound. 
He received the Technical Development Award from the ASJ in 2016, Industrial Standardization Encouragement Awards from Ministry of Economy Trade and Industry (METI) of Japan in 2011, 
the Telecom System Technology Paper Encouragement Award from the Telecommunications Advancement Foundation (TAF) of Japan in 2007. 
He is a member of the ASJ, the IEICE, and the IPSJ.
\end{IEEEbiography}


\begin{thebibliography}{99}
\bibitem{Clavel_2005} C.~Clavel, T.~Ehrette, and G.~Richard
``Events Detection for an Audio-Based Surveillance System,''
{\it In Proc.} of ICME, 2005.
\bibitem{Valenzise2007} G.~Valenzise, L.~Gerosa, M.~Tagliasacchi, F.~Antonacci, and A.~Sarti, 
``Scream and Gunshot Detection and Localization for Audio-Surveillance Systems,''
{\it In Proc.} of AVSS, 2007.
\bibitem{Ntalampiras_2011} S.~Ntalampiras, I.~Potamitis, and N.~Fakotakis
``Probabilistic Novelty Detection for Acoustic Surveillance Under Real-World Conditions,''
IEEE Trans. on Multimedia, pp.713--719, 2011.
\bibitem{Foggia_2016}
P.~Foggia, N.~Petkov, A.~Saggese, N.~Strisciuglio, and M.~Vento,
``Audio Surveillance of Roads: A System for Detecting Anomalous Sounds,''
IEEE Trans. ITS, pp.279--288, 2016.
\bibitem{Coucke_2003} P.~Coucke, B.~De.~Ketelaere, and J.~De.~Baerdemaeker,
``Experimental analysis of the dynamic, mechanical behavior of a chicken egg,''
{\it Journal of Sound and Vibration}, Vol. 266, pp.711--721, 2003.
\bibitem{Chung_2013_pig} Y.~Chung, S.~Oh, J.~Lee, D.~Park, H.~H.~Chang and S.~Kim,
``Automatic Detection and Recognition of Pig Wasting Diseases Using Sound Data in Audio Surveillance Systems,''
Sensors, pp.12929--12942, 2013. 
\bibitem{Yamashita_2006} A.~Yamashita, T.~Hara, and T.~Kaneko,
``Inspection of Visible and Invisible Features of Objects with Image and Sound Signal Processing,''
in \textit{Proceedings of the 2006 IEEE/RSJ International Conference on Intelligent Robots and Systems (IROS2006),} pp. 3837--3842, 2006.
\bibitem{Koizumi_2017_ADS} Y.~Koizumi, S.~Saito, H.~Uematsu, and N.~Harada,
``Optimizing Acoustic Feature Extractor for Anomalous Sound Detection Based on Neyman-Pearson Lemma,''
{\it in Proc.} of EUSIPCO, 2017.





\bibitem{DCASE2017} A.~Mesaros, T.~Heittola, A.~Diment, B.~Elizalde, A.~Shah, E.~Vincent, B.~Raj, and T.~Virtanen,
``DCASE 2017 challenge setup: tasks, datasets and baseline system,''
in Proc. of the Detection and Classification of Acoustic Scenes and Events 2017 Workshop (DCASE2017), pp. 85--92, 2017.
%
\bibitem{rare_01} H.~Lim, J.~Park and Y.~Han,
``Rare Sound Event Detection Using 1D Convolutional Recurrent Neural Networks,''
in Proc. of the Detection and Classification of Acoustic Scenes and Events 2017 Workshop (DCASE2017), 2017.
\bibitem{rare_02} E.~Cakir and T.~Virtanen,
``Convolutional Recurrent Neural Networks for Rare Sound Event Detection,''
in Proc. of the Detection and Classification of Acoustic Scenes and Events 2017 Workshop (DCASE2017), 2017.
%
\bibitem{task1_2016} H.~Eghbal-Zadeh, B.~Lehner, M.~Dorfer, and G.~Widmer,
``CP-JKU Submissions for DCASE-2016: a Hybrid Approach Using Binaural I-Vectors and Deep Convolutional Neural Networks,''
in Proc. of the Detection and Classification of Acoustic Scenes and Events 2016 Workshop (DCASE2016), 2016.
\bibitem{task1_2017} S.~Mun, S.~Park, D.~K.~Han, and H.~Ko,
``Generative Adversarial Network Based Acoustic Scene Training Set Augmentation and Selection Using Svm Hyperplane,''
in Proc. of the Detection and Classification of Acoustic Scenes and Events 2017 Workshop (DCASE2017), 2017.
%
\bibitem{task2_2016} S.~Adavanne, G.~Parascandolo, P.~Pertila, T.~Heittola, and T.~Virtanen,
``Sound Event Detection in Multichannel Audio Using Spatial and Harmonic Features,''
in Proc. of the Detection and Classification of Acoustic Scenes and Events 2016 Workshop (DCASE2016), 2016.
\bibitem{task2_2017} S.~Adavanne, and T.~Virtanen,
``A Report on Sound Event Detection with Different Binaural Features,''
in Proc. of the Detection and Classification of Acoustic Scenes and Events 2017 Workshop (DCASE2017), 2017.
%
\bibitem{tagging} T.~Lidy and A.~Schindler,
``CQT-Based Convolutional Neural Networks for Audio Scene Classification and Domestic Audio Tagging,''
in Proc. of the Detection and Classification of Acoustic Scenes and Events 2016 Workshop (DCASE2016), 2016.


\bibitem{Hodge_2004} V.~J.~Hodge and J.~Austin,
``A Survey of Outlier Detection Methodologies,''
Artificial Intelligence Review, pp 85--126, 2004.
\bibitem{Patcha_2007} A.~Patcha and J.~M.~Park,
``An overview of anomaly detection techniques: Existing solutions and latest technological trends,''
Journal Computer Networks, pp.3448--3470, 2007.
\bibitem{ASD_survey} V.~Chandola, A.~Banerjee, and V.~Kumar
``Anomaly detection: A survey,''
{\it ACM Computing Surveys}, 2009.

\bibitem{Marchi_2015} E.~Marchi, F.~Vesperini, F.~Eyben, S.~Squartini, and B.~Schuller,
``A Novel Approach for Automatic Acoustic Novelty Detection using a Denoising Autoencoder with Bidirectional LSTM Neural Networks,''
{\it In Proc.} of ICASSP, 2015.
\bibitem{Tagawa_2015} T.~Tagawa, Y.~Tadokoro, and T.~Yairi,
``Structured Denoising Autoencoder for Fault Detection and Analysis,''
Proceedings of Machine Learning Research, pp.96--111, 2015.
\bibitem{Marchi_2015_IJCNN} E.~Marchi, F.~Vesperini, F.~Weninger, F.~Eyben, S.~Squartini, and B.~Schuller,
``Non-linear prediction with LSTM recurrent neural networks for acoustic novelty detection,''
{\it In Proc.} of IJCNN, 2015.
\bibitem{Kawaguchi_2017_MLSP} Y.~Kawaguchi and T.~Endo,
``How can we detect anomalies from subsampled audio signals?,''
{\it in Proc.} of MLSP, 2017.
\bibitem{VAE_Anomaly} J.~An and S.~Cho,
``Variational Autoencoder based Anomaly Detection using Reconstruction Probability,''
Technical Report. SNU Data Mining Center, pp.1--18, 2015.
\bibitem{Kawachi_2018} Y.~Kawachi, Y.~Koizumi, and N.~Harada,
``Complementary Set Variational Autoencoder for Supervised Anomaly Detection,''
{\it in Proc.} of ICASSP, 2018. 


\bibitem{GAN} 
I.~J.~Goodfellow, J.~P.~Abadie, M.~Mirza, B.~Xu, D.~W.~Farley, S.~Ozair, A.~Courville, and Y.~Bengio,
``Generative Adversarial Networks,''
{\it In Proc} of NIPS, 2014.
\bibitem{VAEGAN} A.~B.~L.~Larsen, S.~K.~Sonderby, H.~Larochelle, and O.~Winther,
``Autoencoding beyond pixels using a learned similarity metric,''
{\it In Proc.} of ICML, 2016.
\bibitem{GAN_AD} 
T.~Schlegl, P.~Seebock, S.~M.~Waldstein, U.~S.~Erfurth, and G.~Langs,
``Unsupervised Anomaly Detection with Generative Adversarial Networks to Guide Marker Discovery,''
{\it In Proc.} of IPMI, 2017.




\bibitem{Ney_Pear}
J. Neyman and E.~S.~Pearson,
``On the Problem of the Most Efficient Tests of Statistical Hypotheses,''
Phi. Trans. of the Royal Society, 1933.
\bibitem{MostPowerful} G.~Casella and R.~L.~Berger, ``{\it Statistical Inference, section 8.3.2 Most Powerful Test},'' Duxbury Pr, pp.387--393, 2001.




\bibitem{Bradley_1997} A.~P.~Bradley,
``The use of the area under the ROC curve in the evaluation of machine learning algorithms,''
{\it Pattern Recognition}, vol. 30, no. 7, pp. 1145--1159, 1997.
\bibitem{Herschtal_2004} A.~Herschtal and B.~Raskutti,
``Optimising Area Under the ROC Curve Using Gradient Descent,''
{\it In Proc.} of ICML, 2004.
\bibitem{Fujino_2016} A.~Fujino and N.~Ueda,
``A Semi-supervised AUC Optimization Method with Generative Models,''
{\it In Proc.} of ICDM, 2016.




\bibitem{adam} D.~P.~Kingma and J.~L.~Ba,
``Adam: A Method for Stochastic Optimization,''
{\it In Proc.} of ICLR, 2015.
\bibitem{l2_devay} A.~Krogh and J.~A.~Hertz,
``A Simple Weight Decay Can Improve Generalization,''
{\it In Proc.} of NIPS, 1992.
\bibitem{DCASE2016} \url{http://www.cs.tut.fi/sgn/arg/dcase2016/}
\bibitem{dl_url} \url{http://www.cs.tut.fi/sgn/arg/dcase2016/download}




\bibitem{pAUC} S.~D.~Walter
``The partial area under the summary ROC curve,''
{\it Statistics in medicine}, pp.2025--2040, 2005.
\bibitem{Gornitz2013} N.~Gornitz, M.~Kloft, K.~Rieck, and U.~Brefeld,
``Toward Supervised Anomaly Detection,''
Journal of Artificial Intelligence Research, pp.235--262, 2013.


\end{thebibliography}
\end{document}